\documentclass{article}
\usepackage[preprint, nonatbib]{neurips_2026}   % use [final] for camera-ready styling

\usepackage{amsmath}
\usepackage{amstext}
\usepackage{graphicx}
\usepackage{bm}
\usepackage{booktabs}
\usepackage[utf8]{inputenc}
\usepackage[T1]{fontenc}
\usepackage{hyperref}
\hypersetup{
    colorlinks=false,
    pdfborder={0 0 0}
}
\usepackage{url}
\usepackage{microtype}

\usepackage[backend=biber,style=numeric-comp,sorting=none]{biblatex}
\title{Cost-Effective Agent Harnesses for Abstract Reasoning and Generalization on ARC-AGI-1}

\author{%
  Kabir Moghe \\
  Department of Computer Science \\
  Dartmouth College \\
  \texttt{kabir.moghe.26@dartmouth.edu} \\
  \And
  Peter Chin \\
  Department of Computer Science \\
  Dartmouth College \\
  \texttt{peter.chin@dartmouth.edu} \\
}

\begin{document}

\maketitle

\begin{abstract}
		\noindent Recent progress on ARC-AGI-1 from disclosed architectures has come broadly from two regimes: heavy test-time compute over frontier models (evolutionary search, exhaustive sampling, extended chain-of-thought), or benchmark-specific training in which small models are fine-tuned on ARC data, often with task-specialized architectures. We study a third regime: an open-weight model in non-thinking mode (DeepSeek V3.2) under a strict budget, with no ARC-specific fine-tuning. We study what is recoverable through architecture alone, building agentic harnesses that decompose pattern-discovery and program-synthesis stages explicitly. First, we introduce an \textbf{Explorer-Definer Pipeline} that separates pattern discovery from executable transformation synthesis, implemented as a two-stage agent pipeline. Next, we present the \textbf{Reflective Orchestrator}, which augments the pipeline with autonomous exploration of new transformations when previous hypotheses fail on training pairs. On the ARC-AGI-1 public 400-task evaluation set, the pipeline reaches \textbf{57.50\% pass@2 at \$0.25 per task}, and the orchestrator reaches \textbf{67.25\% pass@2 at \$0.62 per task}. Together these architectures lift a 15.50\% one-shot baseline by $\sim$52\ points without benchmark-specific training or heavy test-time compute. Furthermore, the orchestrator-driven lift tests a falsifiable diagnostic the pipeline produces; unbiased pass@k analysis suggests the pipeline is generation-bound, not selection-bound (selection via training-pair accuracy captures $\sim$95\% of the candidate ceiling) and predicts that significant improvement requires broader generation, not better ranking. The orchestrator implements this prediction via adaptive re-exploration and confirms it (unbiased pass@1 lift +9.8\, pp, matching selection-mediated pass@2 lift). An additional pipeline ablation identifies its \texttt{think} tool as a significant component, with removal reducing pass@2 by 5.75\,pp.

\end{abstract}

%----------------------------------------------------------------------------------------
%	1. INTRODUCTION
%----------------------------------------------------------------------------------------

\section{Introduction}
In the years since its inception, ARC has become approachable, but at a cost. Strong reported performance from public work has generally come from one of two regimes. The first relies on heavy test-time compute over frontier large language models (LLMs): evolutionary search over candidate solutions, exhaustive program enumeration, or extended chain-of-thought reasoning, with reported costs ranging from a few dollars to several hundred dollars per task~\cite{berman2024,berman2025,pang2025,greenblatt2024}. The second relies on benchmark-specific training: small models trained from scratch or fine-tuned directly on ARC-distribution data, typically combined with test-time architectures employing fine-tuning on the target task and aggressive data augmentation~\cite{trm2025,architects,arcprize_arcagi1}. Both regimes have successfully advanced leaderboard numbers, but each trades strong performance for substantial compute spent either on significant inference per task or on benchmark-specific training.

\paragraph{Contextualizing the ``no RL'' or fine-tuning condition.}
The scope distinction relevant to this paper is between benchmark-specific training, whether supervised, test-time-finetuned, or RL-based, and general-purpose post-training. The above second-regime systems all fine-tune or train from scratch on ARC training tasks. The model we use --- DeepSeek~V3.2 in non-thinking mode --- has undergone general post-training on broad task mixtures, including significant reinforcement learning components (e.g., RLHF, agentic-capability RL, reasoning-style RL), but has not been specialized to ARC, with the exception of any public data leakage. We hold that general post-training constant across every condition in this paper, so the architectural deltas we report are not confounded by it.

We argue that the gap between cheap one-shot LLM calls ($\sim$16\% on ARC-AGI-1) and ``performant'' leaderboard systems ($\geq$50\%) is not solely a function of model capability or compute scale. Instead, a meaningful share of that gap is recoverable through three distinct architectural levers: \emph{within-call deliberation} (a structured chain-of-thought channel), \emph{across-call decomposition} (separating pattern exploration from transformation synthesis), and \emph{adaptive re-exploration} (allowing the synthesis stage to spawn fresh exploration when its current hypothesis fails, rather than refining based on a fixed prior exploration artifact). We test this hypothesis by comparing four architectures of increasing structural sophistication on the ARC-AGI-1 public evaluation set, all run on the same model.

Our central finding has two parts. First, the Explorer-Definer Pipeline reaches \textbf{57.50\% pass@2 at \$0.25 per task} --- a 42.00-point lift over the one-shot baseline (15.50\%) and a 27.50-point lift over a structured chain-of-thought baseline (30.00\%). Second, the pipeline produces a falsifiable diagnostic: it is \emph{generation-bound, not selection-bound}, meaning that the observed constraint on accuracy is the actual diversity of candidate transformations, not the quality of our selection mechanism. This diagnostic suggests that breaking through the pipeline's ceiling requires broader candidate generation rather than better selection. The Reflective Orchestrator is the architectural test of that insight: it extends the pipeline's definer with tooling to (a) spawn focused mid-loop exploration when its current transformation fails or seems particularly incorrect and (b) to gracefully determine its own loop-exit criteria. The orchestrator architecture reaches \textbf{67.25\% pass@2 at \$0.62 per task}, a \textbf{+9.75-point lift} over the pipeline (paired bootstrap 95\,\% CI $[+6.50, +13.25]$), with a near-identical lift on the unbiased pass@1 estimator that confirms the gain is generation-side rather than selection-mediated. The total architectural lift from one-shot baseline to orchestrator is 52 percentage points on the same model, with no benchmark-specific training or costly, unbounded frontier-model inference.

\paragraph{Public-evaluation caveat.}
We do not report an official ARC Prize leaderboard score. Official verified results are evaluated on the semi-private ARC-AGI-1 holdout, whereas this study uses the public 400-task evaluation set, which is the largest evaluation split available for arbitrary local architecture experiments~\cite{arcprize_policy,arcprize_arcagi1}. This prevents direct leaderboard comparison and introduces the possibility that public-set results overestimate semi-private generalization. To mitigate this, every architecture is evaluated through controlled ablations on the same model and provider, against the same 400 tasks, with the same scoring and cost-accounting code. External system numbers are reported only as contextual reference points for the cost--accuracy regime, not as matched baselines.
\paragraph{Contributions.}
\begin{enumerate}
\item \textbf{Two new agentic architectures occupying a sub-\$1-per-task region of the ARC-AGI-1 cost--accuracy frontier.} The Explorer-Definer Pipeline (57.50\% pass@2 at \$0.25 per task) decomposes ARC into separate pattern-discovery and program-synthesis stages, mediated by compressed natural-language artifacts and verified by training-pair execution. The Reflective Orchestrator (67.25\% pass@2 at \$0.62 per task; $+9.75$\,pp paired) extends the pipeline's definer with a tool to spawn focused mid-loop exploration when its transformation hypothesis fails. Both architectures sit in a region of the cost--accuracy plane sparsely occupied by disclosed architectures.

\item \textbf{A diagnostic-then-test methodology that ties the orchestrator's design to a falsifiable prediction.} We maintain the same 400-task public evaluation set, model, grid rendering, scoring, and cost accounting across conditions. Holding these elements constant, our analysis isolates a likely architectural limitation for the pipeline that appears to bound accuracy even with more exhaustive configurations: the unbiased pass@$k$ estimator~\cite{chen2021codex} applied to the pipeline's $M{=}5$ candidate sets shows that train-score selection captures $\sim$95\% of the achievable candidate ceiling. This diagnostic motivates the idea that pipeline accuracy is constrained by candidate generation, not candidate selection. We implement the orchestrator as a targeted architectural augmentation on the pipeline to test that prediction. Indeed, its near-equal lifts on naive pass@2 ($+9.75$\,pp) and unbiased pass@1 ($+9.81$\,pp) confirm gain is generation-side rather than selection-mediated. 

\item \textbf{Empirical characterization of robustness across architectural lifts.} An exhaustive Pareto sweep of the pipeline across explorer count, temperature, and definer best-of-sampling shows that explorer count is the dominant cost--accuracy lever, best-of-sampling seemingly saturates, and intermediate temperature often dominates. Ablations show that disabling the definer's \texttt{think} scratchpad costs $-5.75$\,pp pass@2, while disabling train-feedback refinement only yields $-0.75$\,pp. Initial cross-model validation on Qwen3-235B (for a matched 99-task subset) across the four architectures follows the same lift trajectory, offering evidence that the harness, not the specific model, drives the lift. Trace-level analysis on the orchestrator shows $\sim$75\% of unique solves utilize adaptive re-exploration, corroborating the design's central hypothesis.

\item \textbf{A methodological discipline for compute-efficiency claims}. Token-normalized per-task cost reporting, explicit pass@$k$ definitions, bootstrap confidence intervals, and paired-bootstrap deltas help make architectural lifts auditable.
\end{enumerate}

\paragraph{Code and reproducibility.}
All code, prompts, harness implementations, and analysis scripts are available at \url{https://github.com/kabirmoghe/cost-effective-arc-harnesses}.

The remainder of this paper is organized as follows. Section~\ref{sec:related} reviews prior work, organized around the mechanisms our architectures use (search vs.\ training regimes on ARC, sampling-based self-consistency, verbal-feedback agents, and tool-mediated reasoning). Section~\ref{sec:methods} details the four architectures, the evaluation protocol, and the cost-accounting conventions. Section~\ref{sec:results} presents controlled architectural deltas, the $N \times t \times M$ Pareto surface, the pipeline ablations, the generation-vs-selection analysis, the pipeline-to-orchestrator M-sweep comparison, cross-model validation on Qwen3-235B, and cost-context comparisons. Sections~\ref{sec:discussion},~\ref{sec:limitations}, and~\ref{sec:futurework} discuss implications, limitations, and proposed extensions. Appendix~\ref{sec:canonical} showcases a canonical task for which the benefit of the orchestrator's adaptive re-exploration component is apparent.

%----------------------------------------------------------------------------------------
%	2. RELATED WORK
%----------------------------------------------------------------------------------------

\section{Related Work}\label{sec:related}

\subsection{ARC-AGI and cost regimes}

The Abstraction and Reasoning Corpus~\cite{chollet2019measure} was introduced as a deliberate counterpoint to benchmarks where high performance can be achieved through scale and pattern matching. Each ARC task is structurally a few-shot learning problem: a small number of input--output grid pairs ($\sim$3 on average across training and evaluation sets) demonstrate a transformation, and the system must apply it to a held-out test input. The benchmark resists shortcut solutions through three design choices: transformations differ across tasks (no transfer between training and evaluation), correctness requires exact cell-level match, and the search space is open-ended (grids up to $30 \times 30$ with up to ten colors yield more candidate outputs than any plausible memorization strategy could cover). ARC-AGI-2~\cite{arcagi2} extends the original benchmark with harder tasks while preserving the same evaluation protocol.

High-performing published systems typically spend substantial compute either during training, during per-task search, or through large test-time candidate generation; we organize them into two regimes.

\paragraph{Frontier-model test-time search.}
Berman~\cite{berman2024,berman2025} demonstrates that frontier LLMs can solve ARC tasks via evolutionary search over candidate solutions. The 2024 system evolves Python transform functions in generations, scoring each on training-set fitness and revising top performers; it reaches 53.6\% on ARC-AGI-1 with Sonnet~3.5 at approximately \$29 per task. The 2025 system retains the evolutionary architecture but evolves natural-language instructions rather than Python, reaching 79.6\% on ARC-AGI-1 with Grok-4 at \$8.42 per task. Greenblatt's earlier system~\cite{greenblatt2024} similarly relies on programmatic synthesis with extensive sampling, at an estimated cost on the order of \$400 per task. Pang~\cite{pang2025} explicitly targets the cost--accuracy frontier within this regime, using a DreamCoder-inspired library of reusable concepts to reach 77.1\% on ARC-AGI-1 at an estimated \$2.56 per task --- the closest published prior work in spirit to this paper, with both framing ARC progress as a Pareto problem rather than being purely accuracy-driven. However, the two sit on opposite sides of a regime boundary: while Pang optimizes efficiency \emph{within} frontier-model evolutionary search, this paper investigates how far one can move by \emph{stepping outside} that regime.

\paragraph{Training-time specialization on ARC.}
A second regime achieves competitive accuracy by training small models specifically for ARC-style tasks, through some combination of supervised training on ARC-distribution data, test-time fine-tuning on the specific task at hand, and aggressive test-time augmentation. The MindsAI~\cite{arcprize_arcagi1} and ARChitects~\cite{architects} competition entries follow this pattern. The Tiny Recursive Model (TRM)~\cite{trm2025} reports 45\% on ARC-AGI-1 with only 7M parameters, framed as evidence that small recursive networks can rival frontier LLMs on abstract reasoning. Subsequent analysis by Roye-Azar et al.~\cite{royeazar2025} complicates this framing: TRM's reported accuracy depends substantially on a 1000-sample test-time augmentation pipeline (responsible for $\sim$10 absolute points), on task-specific identifier embeddings without which accuracy collapses to zero, and on training-time regularization rather than deep iterative inference (single-step prediction captures $\sim$94\% of final accuracy). Their analysis underscores that compute efficiency cannot be reduced to parameter count: total inference cost, including test-time augmentation, remains a meaningful axis.
 
The scope distinction relevant to this paper is between \emph{benchmark-specific} training and \emph{general-purpose} post-training. The systems above all fine-tune or train from scratch on ARC training tasks, effectively ingesting the benchmark's distribution into their weights. We explicitly avoid any benchmark-specific training (whether supervised, test-time-finetuned, or RL-based) such that any model-level specialization on ARC is limited to generalized model post-training; by holding this across all conditions, we ensure deltas are entirely architectural. 

\paragraph{Position of this work.}
The cost--accuracy region below \$1 per task, on an open-weight model in non-thinking mode, with no ARC-specific training, is largely unoccupied by disclosed architectures and reported accuracy above $\sim$30\%. This paper investigates that region directly, with configurations spanning \$0.25 to \$0.62 per task. Our claim is not that this regime achieves a new state of the art; it is that a substantial fraction of ARC-AGI-1 performance is accessible through agent harness engineering alone.

\subsection{Sampling-based selection: self-consistency and best-of-$N$}
Self-consistency~\cite{wang2022selfconsistency} samples $N$ chain-of-thought trajectories from a single model and selects the final answer by majority vote, on the premise that diverse reasoning paths concentrate on the correct symbolic answer more often than on any single incorrect one. Universal Self-Consistency~\cite{chen2023univ_self_consistency} generalizes this to free-form outputs where a majority vote is undefined, replacing the vote with an LLM-mediated consistency judgement.

The pipeline's exploration stage is the closest structural analog: $N$ PatternExplorer agents independently analyze the training pairs through iterative note-taking and ultimately emit free-form natural-language pattern descriptions. The free-form output form puts this closer to Universal Self-Consistency than to vote-based aggregation, and the mechanism is also LLM-mediated: the downstream definer reads all $N$ findings together and weighs them, flexibly attending both to majority patterns and to dissenting signals. Our setup differs from Universal Self-Consistency in two related respects. First, the aggregation is implicit and embedded in a downstream task rather than executed as a standalone selection step: the definer is not selecting one ``best'' explorer finding, it is synthesizing a program that all of them inform. Second, the crisp selection criterion --- exact train-pair execution accuracy --- is applied later, at the synthesis stage, rather than at the explorer-aggregation step where Universal Self-Consistency would apply its LLM-mediated consistency judgement.

The downstream synthesis stage differs substantially in its best-of-$M$ sampling by relying on a deterministic train-pair execution verifier and dedup-by-test-prediction (Section~\ref{sec:pipeline}).

\subsection{Verbal feedback and self-refinement}

Reflexion~\cite{shinn2023reflexion} introduces verbal reinforcement loops in which an agent reflects on failed attempts and revises its approach across episodes, treating failure traces as a non-parametric memory. Language Agent Tree Search~\cite{zhou2024lats} extends best-of-$N$ sampling with explicit tree structure over candidate trajectories. Our pipeline draws on this lineage but differs in one important respect: feedback is consumed \emph{within the same task episode}, not across episodes, with the ``reward'' expressed as structured failure context (input, expected output, predicted output) derived from train-pair execution. ARC's training pairs thus serve as a verifiable per-task feedback signal that the model can perform verbal credit assignment over, without ever updating weights, as with Reflexion. This is also close in spirit to Self-Refine~\cite{madaan2023selfrefine}, adapted to a setting where the correctness signal is exact and program-level rather than a critic LLM's judgement.

The Reflective Orchestrator (Section~\ref{sec:orchestrator}) extends this lineage further. Where Reflexion-style loops alternate generation and reflection at the level of complete trajectories, and where the pipeline's refinement confines revision to the synthesis stage, the orchestrator extends that stage with an explicit mid-loop \emph{re-exploration} tool: when the current transformation hypothesis is failing particularly poorly on train pairs, the orchestrator can branch out of the synthesis frame entirely and spawn fresh pattern-discovery agents conditioned on a structured failure summary. The distinguishing claim is that revision must reach \emph{back into the upstream stage} (exploration) to escape wrong-abstraction failures because refining within a wrong frame produces the ``analysis paralysis'' traces our pipeline data exhibit. In this paper, the branch decision (refine, spawn, or commit) is model-mediated within its agentic loop rather than driven by an explicit confidence estimator or learned branch policy.

\subsection{Tool-mediated reasoning and the \texttt{think} scratchpad}

A separate line of work treats explicit, model-controlled scratchpads as architectural components rather than purely prompt-induced. The \texttt{think} tool~\cite{anthropic_think_tool} is framed as a private deliberation channel for agents that must analyze tool outputs, follow long tool-call chains, and make sequential decisions where mistakes are costly --- distinct from extended-thinking modes that operate before the response begins. The pipeline's PatternExplorer and TransformationDefiner agents both expose a \texttt{think} tool of this form; we report a controlled removal of the definer's \texttt{think} tool in Section~\ref{sec:actonly} that quantifies how much the channel contributes to pipeline accuracy.

\subsection{Neural-guided program synthesis for ARC}

DreamCoder~\cite{ellis2021dreamcoder} treats ARC as a program-induction problem in a domain-specific language, with neural guidance over enumerative search. LARC~\cite{acquaviva2022larc} provides natural-language descriptions of ARC tasks, exploring whether linguistic decomposition aids program synthesis. Our pipeline sits between these traditions: the transformation definer emits a Python function (program synthesis), but the function is generated by a general-purpose LLM rather than searched over a hand-designed DSL, and an upstream exploration phase produces natural-language pattern descriptions that condition the synthesis (linguistic decomposition).

\subsection{Test-time compute scaling}

Snell et al.~\cite{snell2024scaling} characterize the trade-off between model size and test-time sampling, finding that for many tasks, increasing test-time compute on a smaller model outperforms using a larger model with single-pass inference. Process reward models~\cite{lightman2023verify} provide a complementary framing: rather than scoring final outputs, scoring intermediate reasoning steps can enable more targeted allocation of compute. This paper can be read as an architectural complement to both lines: rather than sampling more from a single call or scoring intermediate tokens, the pipeline allocates compute to structurally distinct phases (parallel exploration, focused synthesis, deterministic execution), with verification against training pairs serving as an end-to-end reward signal. The generation-bound diagnostic in Section~\ref{sec:results} sharpens this connection: within our regime, returns from additional definer samples appear to diminish, so from a compute perspective, the productive lever is upstream candidate diversity (explorer count $N$) rather than downstream selection fidelity (definer count $M$).

%----------------------------------------------------------------------------------------
%	3. METHODS
%----------------------------------------------------------------------------------------

\section{Methods}\label{sec:methods}

We evaluate four architectures on the ARC-AGI-1 public evaluation set: a one-shot baseline, a structured chain-of-thought baseline, the Explorer-Definer Pipeline, and the Reflective Orchestrator. Each architecture receives the same input --- a set of training input--output grid pairs $(x_i, y_i)_{i=1}^{k}$ and a test input $x_{\text{test}}$ --- and must produce the corresponding test output $\hat{y}_{\text{test}}$. The pipeline is reported at one canonical operating point ($N{=}5$, $t{=}0.5$, $M{=}5$) and additionally swept across an $N \times t \times M$ grid for Pareto analysis. The orchestrator is reported at the same canonical operating configuration and additionally swept across $M \in \{1, 2, 3, 4, 5\}$ for pipeline-to-orchestrator comparison.

\subsection{Evaluation protocol}\label{sec:protocol}

\paragraph{Dataset.}
ARC-AGI-1 public evaluation set, $n=400$ tasks. The public set is used because the official semi-private leaderboard service is not available for arbitrary local experiments (see Section~\ref{sec:limitations}).

\paragraph{Scoring.}
File-level exact match on the test output grid (every cell must match). pass@$1$ counts a task as solved if the architecture's single submitted output is correct; pass@$2$ counts it as solved if at least one of two submitted outputs is correct. For pipeline and orchestrator conditions, pass@$2$ outputs are the top-2 candidates after dedup-by-test-prediction (Section~\ref{sec:pipeline}).

\paragraph{Model and provider.}
DeepSeek~V3.2 served via AtlasCloud FP8 endpoints. V3.2 supports a thinking-mode toggle; we run it with thinking disabled, holding the model's internal reasoning behavior constant across all architectures. Provider fallback is disabled. Informal probes across providers and quantizations showed insignificant performance impact.

\paragraph{Sampling.}
The one-shot and CoT baselines use temperature~0. The pipeline's canonical configuration uses explorer temperature $t=0.5$ and definer temperature~0; the $N \times t \times M$ sweep varies $t \in \{0.0, 0.5, 1.0\}$ across explorer agents while holding the definer at temperature~0 throughout. The orchestrator uses the same explorer cell (initial fleet of $N{=}5$ explorers at $t{=}0.5$) and its mid-loop spawn tool issues $K{=}2$ focused explorers at $t{=}0.5$ per call.

\paragraph{Subsampling within sweeps.}
The 45 pipeline configurations reported in Section~\ref{sec:surface} are generated from 18 individual runs. The $(N, t)$ axes both require independent runs (nine in total, one per cell of $N \in \{1, 2, 5\} \times t \in \{0.0, 0.5, 1.0\}$) because both parameters shape the explorer fleet's behavior and cannot be recovered post-hoc. The $M$ axis is subsample-able: definer agents within a single $(N, t)$ cell are independent samples from the same distribution, so a single best-of-$M{=}5$ definer pool yields the full $M \in \{1, \dots, 5\}$ curve for that cell by taking the first $k$ candidates for pass@$k$. The full sweep is thus nine explorer-cell runs plus nine matched definer-pool runs, one definer pool per explorer cell. The orchestrator's $M$-sweep follows the same subsampling protocol within its single $(N{=}5, t{=}0.5)$ explorer cell. This protocol both reduces compute and controls for explorer-side run-to-run variance across the $M$ curves: for fixed $(N, t)$, every $M$ value traces the same set of explorer findings rather than independently drawn ones.

\paragraph{Cost accounting.}
Per-task dollar cost is computed from raw input/output token counts at AtlasCloud FP8 list rates (\$0.30/M prompt, \$0.38/M completion). Costs are reported as the cost of the \emph{architecture} (every explorer call, definer call, retry, and spawn), not the cheapest available deployment; batch-inference discounts, alternative quantization, or self-hosting could reduce absolute costs without changing the relative ordering. Parallelism affects wall-clock but not dollar cost.

\paragraph{Execution sandbox.}
Definer-generated Python is executed in a restricted subprocess with a 10-second per-call timeout, a small allowed-import list (\texttt{numpy}, standard library), and no network access. Runtime errors and timeouts are caught and recorded as a failed candidate (training-pair accuracy zero); they do not abort the per-task pipeline.

\paragraph{pass@$k$ estimator.}
For diagnostic purposes (Section~\ref{sec:passk}), we report both naive pass@$k$ (the empirical fraction of tasks for which the top-$k$ selected candidates contain a correct one) and the unbiased Codex estimator~\cite{chen2021codex}:
\begin{equation*}
\widehat{\mathrm{pass@}k} \;=\; 1 - \frac{\binom{n - c}{k}}{\binom{n}{k}},
\end{equation*}
where $n$ is the number of candidates generated for a task, $c$ is the number of correct candidates among them, and $k$ is the number of attempts allowed. The unbiased estimator characterizes what fraction of tasks \emph{any} $k$ randomly chosen candidates would solve; the gap between it and naive pass@$k$ isolates the contribution of the selection policy.

\paragraph{Uncertainty.}
Bootstrap 95\,\% confidence intervals on accuracy are computed by resampling the 400-task per-task binary correctness vector 10{,}000 times. Paired architectural deltas are reported with matched-task paired-bootstrap CIs; significance is marked at the 95\,\% level.

\subsection{Structured one-shot baseline}

Our ``one-shot'' (in the sense of requiring output in a single pass) baseline establishes a lower bound by making a single LLM call per task with no explicit reasoning scaffolding. Training pairs are serialized into the prompt as ASCII grids with row and column coordinate headers (the same rendering used by CoT and pipeline/orchestrator agents), and the model is instructed to return the predicted output grid via a structured JSON response containing a single \texttt{output} field as a 2D list of integers. No intermediate artifacts, tool calls, or self-correction opportunities are provided.

This baseline serves two purposes. First, it provides a floor against which chain-of-thought and agentic overhead can be justified. Second, it quantifies how much ARC performance is already latent in the pretrained model's weights without architectural intervention (a coarse control for benchmark exposure during training).

\subsection{Chain-of-thought baseline}

The CoT baseline extends the one-shot architecture with a single structural modification: the JSON response schema gains a \texttt{reasoning} field placed before the \texttt{output} field, and the system prompt instructs the model to use it. The \texttt{reasoning} field contains free-flowing natural-language analysis of the observed patterns before the model commits to the \texttt{output} grid.

\begin{verbatim}
{
  "reasoning": "<analysis of patterns>",
  "output": [[...], [...], ...]
}
\end{verbatim}

The ordering constraint is the operative mechanism: requiring externalized reasoning before grid generation forces dedicated token budget for pattern analysis rather than jumping straight to an output grid. This is functionally analogous to chain-of-thought prompting~\cite{wei2022cot} but structurally enforced through the output schema. It is distinct from native reasoning channels (e.g., DeepSeek-R1's \texttt{<think>} block, OpenAI o1-series internal tokens), which provide a separate hidden budget that the model has typically been post-trained via RL to use for deliberation. Our setting excludes such mechanisms.

This condition differs from the one-shot condition in exactly two respects: the CoT prompting layer and the \texttt{reasoning} field added to the (otherwise shared) JSON output schema. Model, grid serialization, task framing, JSON output enforcement, and inference parameters are held identical. This isolates the contribution of structurally enforced externalized reasoning under single-call inference.

\subsection{Explorer-Definer Pipeline}\label{sec:pipeline}

The Explorer-Definer Pipeline is the first of our two agentic architectures. It decomposes ARC task-solving into two sequential phases --- \emph{pattern exploration} and \emph{transformation definition} --- mediated by compressed natural-language artifacts.

\subsubsection{Phase~1: parallel exploration}

A fleet of $N$ PatternExplorer agents independently analyzes the training pairs. Each explorer receives the full set of training pairs and is equipped with two tools:

\begin{itemize}
\item \textbf{\texttt{think(thought)}}: a scratchpad tool that lets the agent record intermediate observations, hypotheses, and reasoning traces without committing to any output. \texttt{think} calls are invisible to other agents and to downstream stages.
\item \textbf{\texttt{note\_pattern(pattern)}}: a structured logging tool that records a candidate pattern observation with a field for the pattern description and any other notes the agent determines as substantive (e.g., grid coordinates or example indices).
\end{itemize}

\begin{figure}[htbp]
    \centering
    \includegraphics[width=0.95\textwidth]{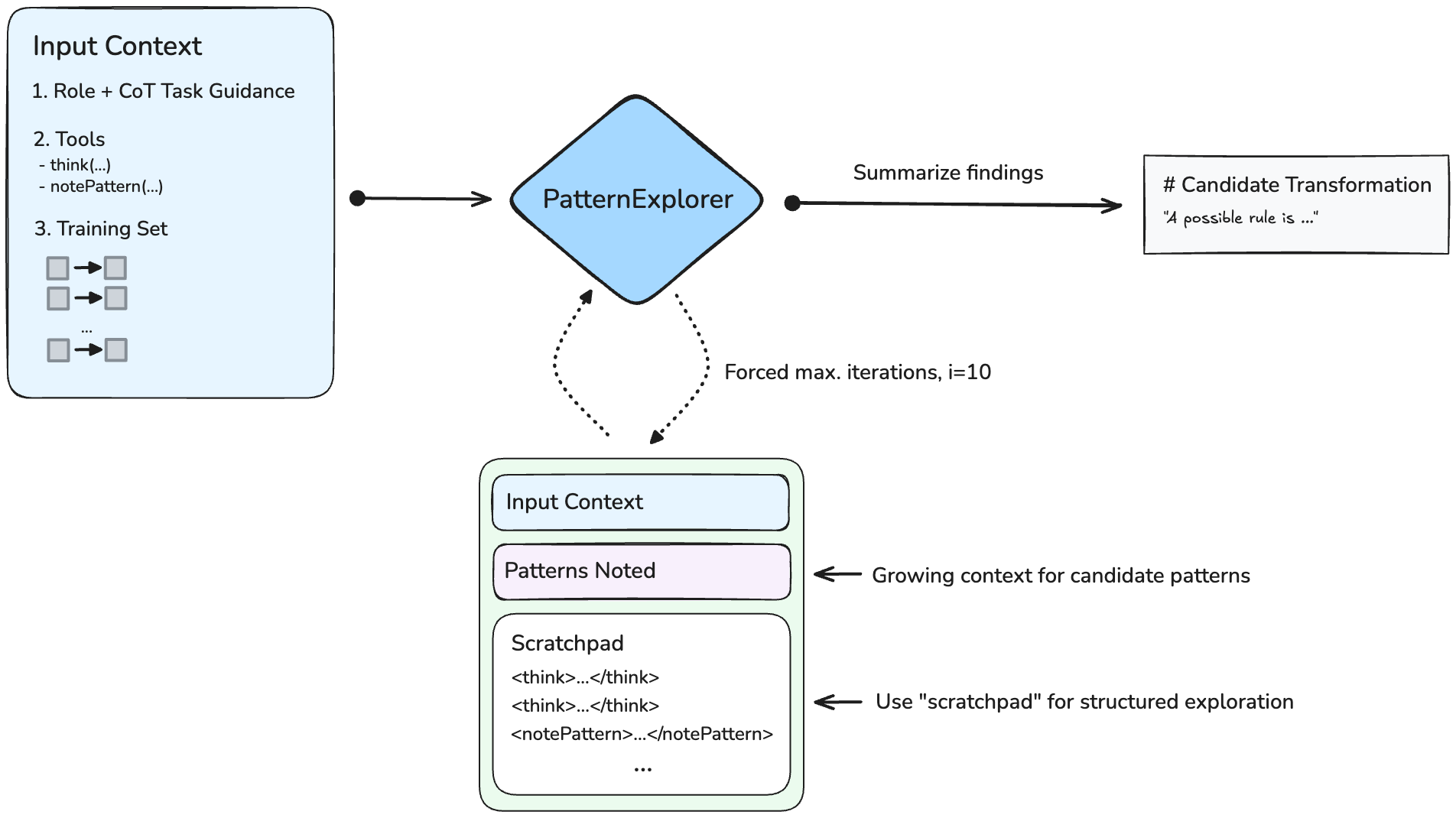}
    \caption{PatternExplorer agent.}
    \label{fig:explorer}
\end{figure}

Explorers are given a generous per-agent token budget ($\sim$100k tokens) and are instructed to be exhaustive: examine symmetries, color mappings, spatial transformations, object detection, counting regularities, and any other structural properties of the input--output pairs. At the end of exploration, each agent is forced to emit a synthesis output containing a structured markdown summary of its findings, including a high-level rule description, skeleton procedure, observations, and self-confidence assessment (e.g., ``low''). 

The $N$ synthesis outputs are compressed into a single artifact of approximately $4k$ tokens per explorer, yielding a combined exploration summary of roughly $4Nk$ tokens. This compression step is significant from cost, context-window, and efficiency standpoints: it distills potentially hundreds of thousands of tokens of exploratory reasoning into a dense signal that fits within a single downstream context window without dominating it.

\subsubsection{Phase~2: transformation definition}

A TransformationDefiner agent receives three inputs: (1)~a system prompt specifying the transformation-definition task, (2)~the original training pairs, and (3)~the compressed exploration summaries. The definer is equipped with three tools:

\begin{itemize}
\item \textbf{\texttt{think(thought)}}: identical to the explorer \texttt{think} tool, for deliberation over evidence.
\item \textbf{\texttt{define\_transformation(transformation\_summary, reasoning, code)}}: a structured output tool with fields for a natural-language description of the transformation, reasoning, and a Python function that implements the transformation as executable code.
\item \textbf{\texttt{submit\_refined\_transformation(what\_changed, code)}}: a structured output tool for refining a previous transformation, fitted with a field to describe the refinement and a field containing the actual refined Python transformation.
\end{itemize}

\begin{figure}[htbp]
    \centering
    \includegraphics[width=0.95\textwidth]{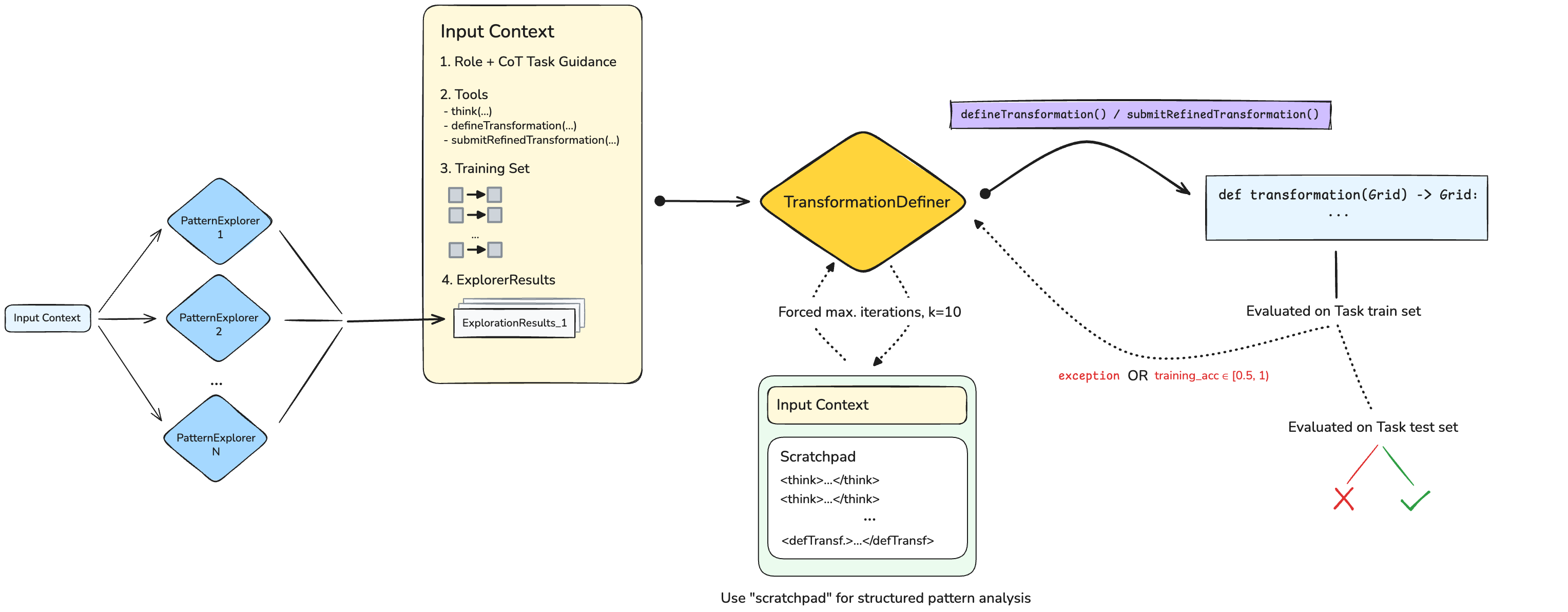}
    \caption{Explorer-Definer Pipeline with a single TransformationDefiner agent.}
    \label{fig:definer}
\end{figure}

The Python transformation function takes an input grid (list of lists of integers) and returns the predicted output grid. By requiring executable code rather than a direct grid output, the pipeline enforces algorithmic generalization: the model must express the transformation as a robust program, not merely pattern-match on the training pairs. Each candidate program is run on every training input and scored by exact-match accuracy on the corresponding training output; programs that fail to execute (errors, timeouts) score zero after the definer has had the opportunity to debug and resubmit. Additionally, transformation-as-code enables producing a heuristic for deterministic selection, detailed as follows.

\paragraph{Selection by dedup-on-test-prediction.}
Among the $M$ candidates produced for a task, we form the pass@$k$ submission by (a) ranking on training-set exact-match score, taking inspiration from Berman's selection mechanism ~\cite{berman2024}, (b) then deduplicating by the predicted test grid: multiple definers that emit lexically distinct programs but produce the same test output are merged into a single candidate, with the highest-ranked retained. Ties on training-set score are broken by agent index (effectively first-emitted). The top-$k$ surviving candidates are the pass@$k$ submission. This deduplication step is not cosmetic: definers frequently converge on the same wrong rule (a common failure mode discussed in Section~\ref{sec:results}), and treating those convergent failures as a single vote prevents crowding out a dissenting correct candidate. The mechanism functions as implicit diverse voting.

\paragraph{Train-feedback refinement.}

When a candidate produces a program that passes at least half but not all of the training pairs, the definer is re-invoked with a structured failure summary (the failing input, the expected output, and the program's actual output) as additional context. The original Phase 2 definer system prompt makes no mention of refinement, but when refinement begins, an augmented prompt is placed at the beginning of context outlining refinement behavior. One refinement pass is allowed per definer. Refinement is skipped if the initial definition reaches 100\% training accuracy. Otherwise, refinement exits when refinement cap is reached, or the per-definer step budget is exhausted. We determined the 50\% training-pair threshold for refinement via probes into the distribution of training accuracy vs. successful refinement, though this has not been rigorously tuned. Intuitively, this design takes advantage of the fact that a transformation that performs \textit{moderately} well on a task is more likely to be somewhat correct and therefore refinable.

\subsection{Reflective Orchestrator}\label{sec:orchestrator}

The Reflective Orchestrator is the second of our agentic architectures and the architectural response to the pipeline's generation-bound diagnostic (Section~\ref{sec:passk}).

\paragraph{Motivation.}
The pipeline's definer faces a structural problem visible in trace inspection of its failures. The definer receives a one-time batch of upstream explorer findings before it begins and has \emph{no mechanism to revisit exploration}. When the explorers miss the right abstraction --- a recurring failure mode visible in trace inspection of the pipeline, quantified in Section~\ref{sec:behavior} --- the definer can correctly diagnose from train feedback that its current hypothesis is wrong, but it can only refine within the same wrong frame. Refinement, by design, edits the current program; it cannot reach back into the upstream stage and request a different conceptual frame. The behavioral consequence is a recurring trace pattern in which the definer correctly identifies what is failing and then either iterates on an arithmetic hack that fits training pairs but not the test pair, or stalls (``analysis paralysis''), producing increasingly elaborate edits that do not converge.

Candidate pattern discovery is the explorer's intended specialization, not the definer's, and the pipeline's directed handoff prevents the definer from routing back to it. The orchestrator's design insight is to avoid failures from flawed upstream inputs by enabling an agent-driven loop: give the definer a tool to spawn a smaller fleet of fresh, focused exploration mid-loop, so when its current transformation is failing, it can request new abstractions rather than getting stuck refining an incorrect one. This spawning mechanism functions as the \emph{relief valve} for wrong-abstraction failures.

\paragraph{Architecture.}
A single agentic loop replaces the pipeline's directed Phase 2 entirely. Phase 1 runs identically and seeds an initial set of pattern findings, after which an augmented definer-as-orchestrator agent runs with five tools available throughout:

\begin{itemize}
\item \texttt{think(thought)}: private scratchpad, identical to the pipeline definer's.
\item \texttt{define\_transformation(\dots)}: first commit on a transformation; structurally analogous to the pipeline's \texttt{define\_transformation} tool.
\item \texttt{submit\_refined\_transformation(\dots)}: a refined transformation after seeing train feedback on a prior \texttt{define\_transformation} attempt.
\item \texttt{explore\_new\_patterns(guidance)}: spawn $K{=}2$ focused \texttt{PatternExplorer} agents conditioned on a structured guidance string from the orchestrator. Their resulting pattern documents are appended to the context the orchestrator sees on subsequent steps.
\item \texttt{done(reason)}: agent-mediated termination, guided to call when progress stalls.
\end{itemize}

\begin{figure}[htbp]
    \centering
    \includegraphics[width=0.95\textwidth]{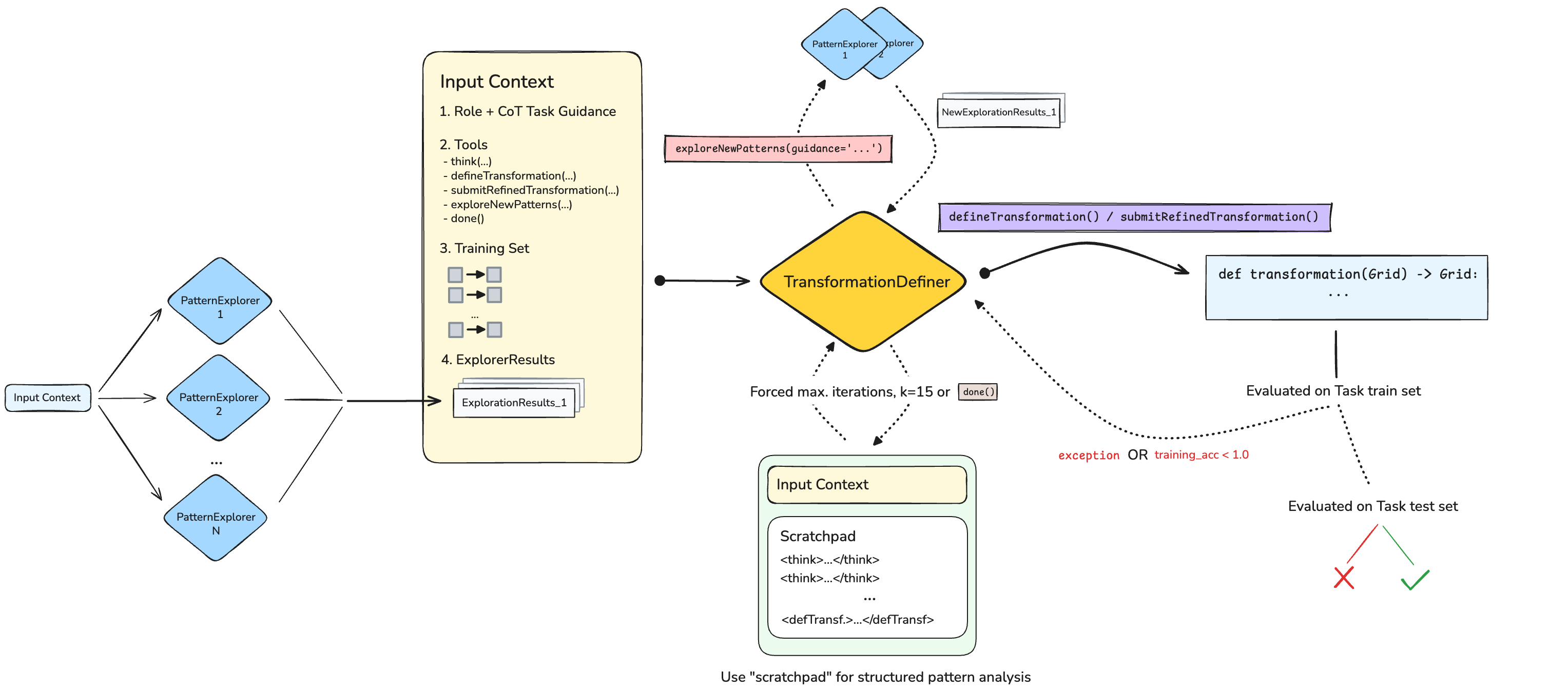}
    \caption{TransformationDefiner as a Reflective Orchestrator.}
    \label{fig:orchestrator}
\end{figure}

The orchestrator exits on one of three conditions: (1) a structural exit when train-set accuracy reaches 100\% on a candidate program, (2) an agent-mediated \texttt{done} call, or (3) one of three fuses (\texttt{MAX\_ITERATIONS}{=}15, \texttt{MAX\_CONSECUTIVE\_EXEC\_ERRORS}{=}5, \texttt{MAX\_SPAWN\_CALLS}{=}3). The branch decision --- continue refining, spawn fresh exploration, or commit and finish --- is model-mediated rather than driven by an explicit confidence estimator, informed at each step by the structured train-pair diffs from the previous attempt.

\paragraph{The $K{=}2$ spawn fan-out.}
The choice of $K{=}2$ explorers per spawn is motivated by two empirical observations from the pipeline sweep (Section~\ref{sec:surface}). First, the largest marginal pass@2 gain along the explorer-count axis is the step from $N{=}1$ to $N{=}2$; subsequent additions appear to diminish. Second, the orchestrator may spawn up to three times within a single task, which compounds quickly: at $K{=}5$ a single spawn would equal the cost of an entire pipeline exploration phase. $K{=}2$ sits at the high-value end of the $N$ curve while keeping per-spawn cost bounded, so spawn behaves as a targeted re-exploration rather than as a second full pipeline invocation. Additionally, the step budget per spawn explorer is intentionally smaller than the Phase 1 explorers to encourage targeted iteration.

\paragraph{Negative-leaning guidance.}
The \texttt{guidance} string the orchestrator passes to spawned explorers is, by prompt design, primarily \emph{negative}: confirmed dead-ends, rule families that the orchestrator has ruled out from its own attempts, and asymmetries in the training pairs that prior hypotheses failed to account for. This design choice reflects an information-theoretic observation: after seeing its own attempts fail, the orchestrator's highest-information signal about the transformation space is which regions of that space are confirmed empty. Open-ended re-exploration would risk reproducing the explorers' original (and known-wrong) trajectory; constraint-shaped exploration redirects attention to the unexamined regions. Appendix~\ref{sec:canonical} highlights how this design choice can improve convergence.

The orchestrator's selection logic is identical to the pipeline's: orchestrator definers run in parallel, their candidates are dedup-ranked by training-set score, and pass@$k$ is the top-$k$ surviving candidates after dedup-by-test-prediction.

\subsection{Ablations and sweeps}\label{sec:ablations}

\paragraph{$N \times t \times M$ Pareto sweep on the pipeline.}
We sweep all combinations of $N \in \{1, 2, 5\}$, explorer temperature $t \in \{0.0, 0.5, 1.0\}$, and definer best-of-$M$ with $M \in \{1, 2, 3, 4, 5\}$; 45 configurations in total, generated via the subsampling protocol described in Section~\ref{sec:protocol}. This characterizes the cost--accuracy frontier of the pipeline itself and distinguishes the contributions of exploration breadth ($N$), exploration diversity ($t$), and selection from a sampled candidate pool ($M$).

\paragraph{Pipeline capability ablations.}
Two architectural features of the canonical pipeline are ablated by removal: (a) the definer's \texttt{think} tool (\emph{act-only}), in which the model must reason within the structured tool fields rather than via a separate scratchpad; and (b) train-feedback refinement (\emph{refinement-off}), in which the definer commits its first program with no opportunity to refine from train-pair failures. Refinement-off is recovered from the canonical pipeline run by subsampling first-attempt outputs (Section~\ref{sec:protocol}), exploiting the fact that the definer's original Phase 2 prompt is refinement-unaware, so first-attempt programs are identically distributed between refinement-on and refinement-off runs; act-only required a dedicated run.

\paragraph{Pipeline-to-Orchestrator M-sweep.}
We additionally report the orchestrator across $M \in \{1, 2, 3, 4, 5\}$ via the same definer-subsampling protocol, holding the upstream explorer cell ($N{=}5$, $t{=}0.5$) fixed. This isolates the orchestrator's lift over the pipeline at matched definer budget.
%----------------------------------------------------------------------------------------
%	4. RESULTS
%----------------------------------------------------------------------------------------

\section{Results}\label{sec:results}

\subsection{Overview}

The canonical Explorer-Definer Pipeline ($N{=}5$, $t{=}0.5$, $M{=}5$) reaches \textbf{57.50\% pass@2 at \$0.25 per task}, and the Reflective Orchestrator ($M{=}5$ on the same explorer cell) reaches \textbf{67.25\% pass@2 at \$0.62 per task}. Relative to the one-shot baseline (15.50\% at \$0.002 per task), the pipeline is a 42.00-point lift at $\sim$125$\times$ higher per-task cost and the orchestrator is a 51.75-point lift at $\sim$310$\times$. Relative to the pipeline, the orchestrator is a \textbf{+9.75-point pass@2 lift} at $\sim$2.5$\times$ cost; the paired bootstrap on this delta has 100\% of replicates positive and excludes zero at the 95\,\% level. All three architectural levers --- within-call deliberation (CoT), across-call decomposition (pipeline), and adaptive re-exploration (orchestrator) --- contribute statistically significant, paired-bootstrap-confirmed lifts.

\subsection{Per-architecture results}

Table~\ref{tab:results} reports file-level accuracy and per-task cost for the four primary architectures; pipeline ablations follow separately in Section~\ref{sec:actonly}.

\begin{table}[htbp]
\centering
\caption{Accuracy and cost across the four primary architectures. DeepSeek~V3.2 served via AtlasCloud FP8, ARC-AGI-1 public evaluation set ($n=400$). Single-call baselines at temperature~0; pipeline and orchestrator use explorer $t=0.5$, definer/orchestrator $t=0$. Cost at AtlasCloud FP8 list rates (\$0.30/M prompt, \$0.38/M completion). Pipeline ablations are reported separately in Table~\ref{tab:ablations}.}
\label{tab:results}
\small
\begin{tabular}{lcccc}
\toprule
Architecture & Metric & Accuracy & 95\,\% CI & \$/task \\
\midrule
Baseline (no-CoT)                                  & pass@1 & 15.50\,\%           & [12.00,\; 19.25]  & \$0.002  \\
CoT                                                & pass@1 & 30.00\,\%           & [25.50,\; 34.50]  & \$0.004  \\
Pipeline ($N{=}5,\,t{=}0.5,\,M{=}5$)               & pass@1 & 54.75\,\%           & [49.75,\; 59.50]  & \$0.25   \\
Pipeline ($N{=}5,\,t{=}0.5,\,M{=}5$)               & pass@2 & 57.50\,\%           & [52.75,\; 62.50]  & \$0.25   \\
\textbf{Reflective Orchestrator ($M{=}5$)}         & \textbf{pass@2} & \textbf{67.25\,\%}  & \textbf{[62.75,\; 71.75]}  & \textbf{\$0.62}   \\
\bottomrule
\end{tabular}
\end{table}

The structured CoT baseline reaches 30.00\% pass@1 (15.50\% $\to$ 30.00\%, essentially a doubling of the one-shot baseline) at slightly more than double the cost, demonstrating that within-call deliberation is a real, measurable lever on this model. The pipeline then nearly doubles accuracy again (30.00\% $\to$ 54.75\% pass@1, 57.50\% pass@2), demonstrating that across-call decomposition is a separate and larger lever. The orchestrator adds a further $+9.75$\,pp pass@2 over the pipeline at the same explorer cell, demonstrating that adaptive re-exploration is yet another distinct lever that the pipeline misses.

\subsection{Paired architectural comparisons}
\label{sec:paired}

Table~\ref{tab:deltas} reports matched-task paired-bootstrap deltas for each architectural transition. A starred entry indicates that the 95\,\% CI excludes zero.

\begin{table}[htbp]
\centering
\caption{Paired-bootstrap accuracy deltas (matched-task, 95\,\% CI; ${}^*$ CI excludes 0).}
\label{tab:deltas}
\small
\begin{tabular}{lcc}
\toprule
Transition & $\Delta$ (pp) & 95\,\% CI \\
\midrule
Baseline $\to$ CoT                                    & $+14.50^*$ & $[+9.75,\; +19.25]$  \\
CoT $\to$ Pipeline (pass@1)                           & $+24.75^*$ & $[+20.00,\; +29.50]$ \\
CoT $\to$ Pipeline (pass@2)                           & $+27.50^*$ & $[+23.00,\; +32.00]$ \\
Pipeline pass@1 $\to$ pass@2                          & $+2.75^*$  & $[+1.25,\; +4.50]$   \\
\textbf{Pipeline $\to$ Reflective Orchestrator (pass@2)} & $\bm{+9.75^*}$ & $\bm{[+6.50,\; +13.25]}$ \\
Baseline $\to$ Reflective Orchestrator (pass@2)       & $+51.75^*$ & $[+46.50,\; +57.00]$ \\
\bottomrule
\end{tabular}
\end{table}

Deltas first confirm structured CoT produces a substantial, statistically significant lift on this model: $+14.50$\,pp over the unscaffolded baseline. Second, the transition from the strongest single-call architecture (CoT) to the pipeline produces a $+27.50$\,pp pass@2 lift. This is comparable in magnitude to the total accuracy of the CoT baseline itself and attributable largely to architectural decomposition. Third, the pipeline-to-orchestrator transition produces a $+9.75$\,pp pass@2 lift whose paired bootstrap is materially tighter than naive subtraction of the marginal CIs would suggest (half-width $\pm 3.38$\,pp paired vs.\ $\pm 9.25$\,pp from independent CI half-widths summed), since pairing controls for task-level difficulty. The CI excludes zero and 100\,\% of bootstrap replicates have $\Delta > 0$.

\subsection{$N \times t \times M$ Pareto surface (Pipeline)}\label{sec:surface}

\begin{figure}[htbp]
    \centering
    \includegraphics[width=\textwidth]{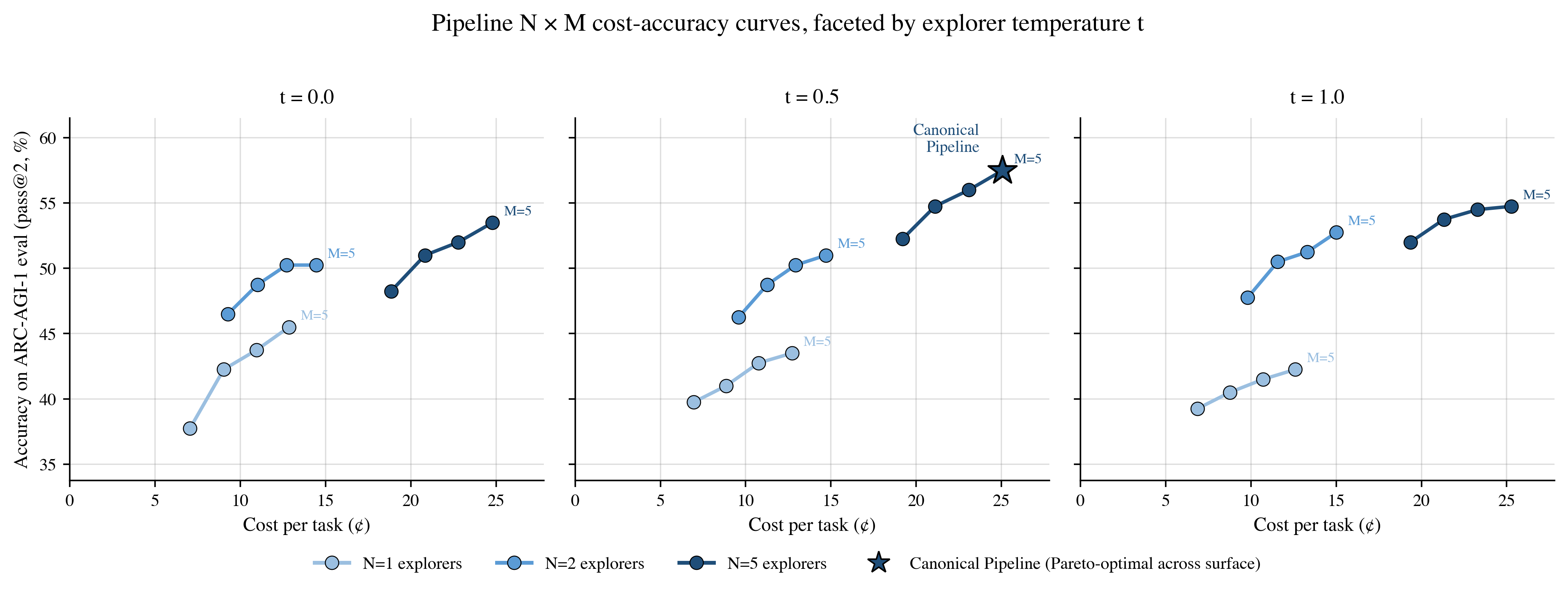}
    \caption{$N \times t \times M$ Pareto surface for the Explorer-Definer Pipeline. DeepSeek~V3.2 served via AtlasCloud FP8, ARC-AGI-1 public evaluation set ($n=400$), refinement enabled. Marker shade: explorer count $N \in \{1, 2, 5\}$. Facet: explorer temperature $t \in \{0.0, 0.5, 1.0\}$. Label: definer best-of-$M$ samples, $M \in \{2, 3, 4, 5\}$. The canonical configuration ($N{=}5,\,t{=}0.5,\,M{=}5$) is starred.}
    \label{fig:surface}
\end{figure}

Figure~\ref{fig:surface} plots the 45 pipeline configurations on the cost--pass@2 plane, faceted by explorer temperature. Three structural facts emerge.

\paragraph{Explorer count $N$ is the dominant lever.}
For each $N$, accuracy appears to saturate: $N{=}1$ never exceeds 45.5\% pass@2 even at $M{=}5$; $N{=}2$ caps at 52.75\%; only $N{=}5$ reaches the 57.50\% headline. Moving along the frontier from low to high cost is dominated by increasing $N$, not $M$. This pattern is the surface-level counterpart of the generation-bound diagnostic in Section~\ref{sec:passk}: candidate diversity (driven upstream by $N$) buys more accuracy than candidate count (driven downstream by $M$).

\paragraph{Best-of-$M$ shows diminishing returns within each $N$.}
Holding $N$ and $t$ fixed, marginal pass@2 from each additional definer sample shrinks: the largest gains come from $M{=}1 \to M{=}2$, and additions past $M{=}3$ contribute only a few percentage points each. The canonical configuration uses $M{=}5$ because it sits on the frontier at the operating cost we target, not because $M$ extrapolates indefinitely.\footnote{This configuration was selected by inspecting the Pareto surface on the same 400-task ARC-AGI-1 public evaluation set used for downstream comparisons. See Section~\ref{sec:limitations}, paragraph \emph{Configuration selection on the evaluation set}, for the methodological caveat and the mitigating considerations specific to this work.} The $M{=}1 \to M{=}2$ marginal also motivates the orchestrator's $K{=}2$ spawn (Section~\ref{sec:orchestrator}).

\paragraph{Intermediate explorer temperature ($t{=}0.5$) dominates.}
At $N{=}5$, the $t{=}0.5$ configurations form the Pareto frontier; $t{=}0.0$ underperforms (less candidate diversity) and $t{=}1.0$ underperforms (degraded individual-trajectory quality). The effect is consistent though small in absolute magnitude.

\subsection{Pipeline ablations: \texttt{think} and refinement}\label{sec:actonly}

Two capability removals on the canonical pipeline isolate which architectural choices are load-bearing. Table~\ref{tab:ablations} reports each removal's effect on pass@2 and cost.

\begin{table}[htbp]
\centering
\caption{Pipeline capability removals. Each row is the canonical pipeline configuration ($N{=}5,\,t{=}0.5,\,M{=}5$) with exactly one feature disabled. Pass@2 is file-level on the ARC-AGI-1 public evaluation set ($n=400$); 95\,\% CIs are matched-task paired bootstrap (B=10{,}000) versus the canonical pipeline in the same (removal) direction as $\Delta$, ${}^*$ excludes zero. Refinement-off is recovered by subsampling first-attempt outputs from the canonical run, exploiting the fact that the initial Phase 2 prompt is refinement-unaware; cost is upper-bounded by the canonical run's \$0.25, with the gap depending on how often refinement triggers per task. Act-only required a dedicated run.}
\label{tab:ablations}
\small
\begin{tabular}{lcccc}
\toprule
Removed capability & pass@2 & $\Delta$ vs. canonical & 95\,\% CI & \$/task \\
\midrule
\emph{(none --- Canonical Pipeline)}        & 57.50\,\% & ---       & ---                  & \$0.25                \\
\texttt{think} tool (act-only)              & 51.75\,\% & $-5.75^*$ & $[-9.00,\;-2.50]$    & \$0.17 ($-32\%$)      \\
Refinement                                  & 56.75\,\% & $-0.75$   & $[-2.00,\;+0.25]$    & $\leq$\,\$0.25        \\
\bottomrule
\end{tabular}
\end{table}

The two removals isolate very different things. The \texttt{think} scratchpad is a measurable component of pipeline performance: removing it costs $\sim$6 percentage points of pass@2, a footprint comparable in magnitude to halving the explorer fleet ($N{=}5 \to N{=}2$) along the Pareto surface, with a paired-bootstrap CI that excludes zero comfortably. The ablation does not separate whether the contribution comes from additional deliberation tokens, the structured separation of a private scratchpad from output channels, or both; we return to this in Section~\ref{sec:discussion}. What it does establish is that the structured tool fields available in the act-only configuration cannot substitute for the scratchpad \texttt{think} provides, consistent with framings of the \texttt{think} tool as a deliberation channel for agents that must analyze tool outputs, follow long tool-call chains, and make sequential decisions where mistakes are costly~\cite{anthropic_think_tool}.

Refinement, by contrast, contributes a much smaller $-0.75$\,pp on removal, with a paired-bootstrap CI that straddles zero ($[-2.00,\;+0.25]$). The removal-$\Delta$ is negative in 95\,\% of bootstrap replicates (consistent with refinement helping) but the small magnitude offers directional evidence rather than a well-resolved effect. Most of refinement's per-definer benefit is plausibly absorbed by the selection layer: refinement helps only when it converts a wrong-train candidate into a right-train one, and the dedup-by-prediction selection already prevents convergent wrong candidates from crowding out a single correct dissenter. 

The two ablations together identify agentic decomposition and the \texttt{think} scratchpad as load-bearing components of the pipeline; refinement is a small additive boost. Both components were thus preserved in the orchestrator's architecture.

\subsection{Pipeline vs.\ orchestrator across $M$}\label{sec:msweep}

\begin{figure}[htbp]
    \centering
    \includegraphics[width=0.85\textwidth]{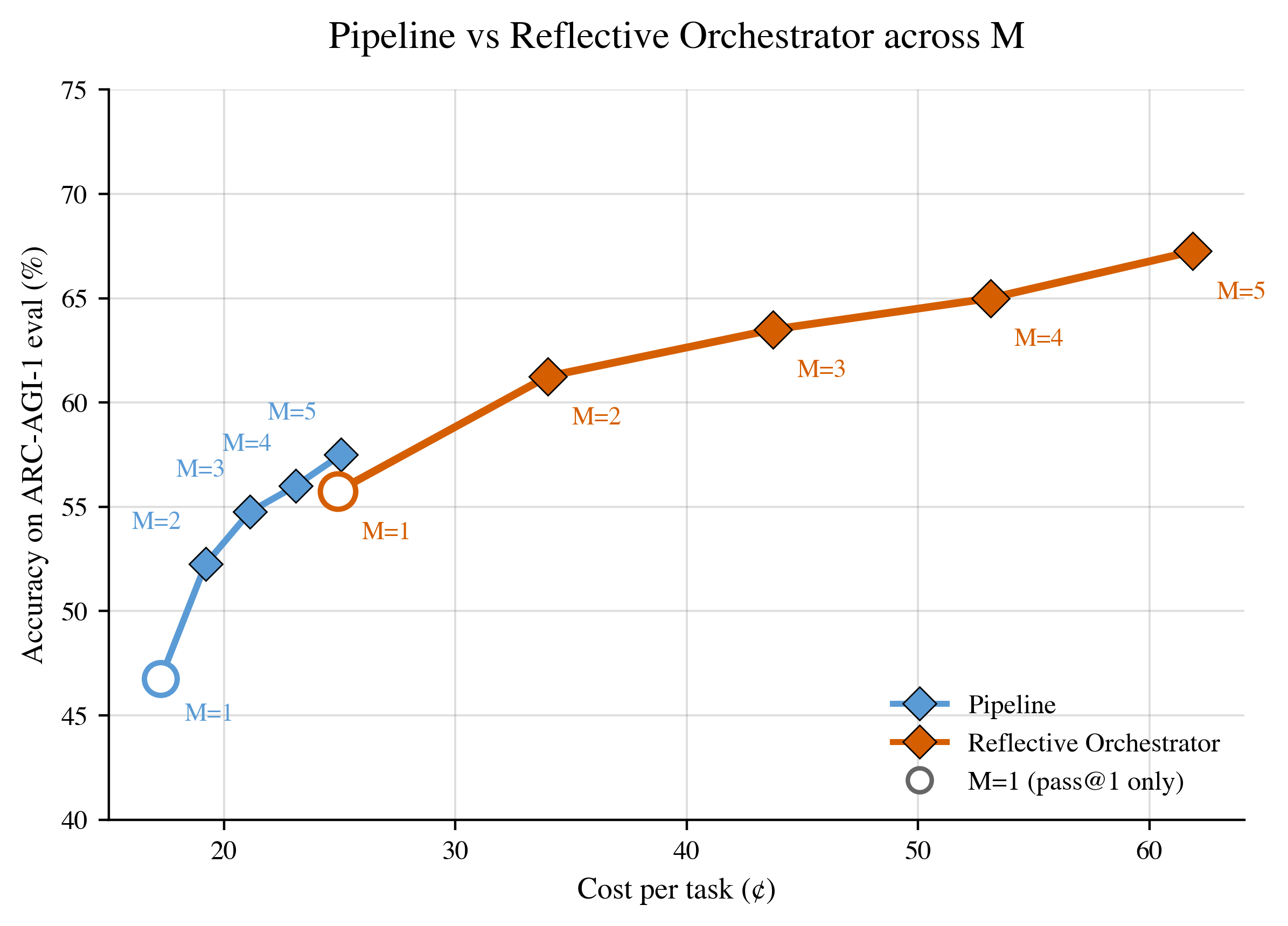}
    \caption{Explorer-Definer Pipeline and Reflective Orchestrator across $M \in \{1,\dots,5\}$ at fixed explorer cell ($N{=}5$, $t{=}0.5$). Hollow markers denote $M{=}1$ (no pass@2 available; pass@1 plotted). The orchestrator Pareto-dominates the pipeline at every $M \geq 2$, with the gap widening as $M$ grows.}
    \label{fig:msweep}
\end{figure}

Figure~\ref{fig:msweep} plots the pipeline and the Reflective Orchestrator across $M \in \{1, 2, 3, 4, 5\}$ on the same explorer cell ($N{=}5$, $t{=}0.5$). The orchestrator Pareto-dominates the pipeline at every $M \geq 2$: at matched $M$, accuracy is higher by 5--10 points; at matched cost, the orchestrator delivers comparable or higher accuracy past the pipeline's likely $M{=}5$ saturation point.

The shapes of the two curves are informative beyond the headline gap. The pipeline appears to saturate between $M{=}3$ and $M{=}5$: each additional definer adds less than a percentage point of pass@2, consistent with the candidate-distribution bimodality the generation-bound diagnostic identifies (Section~\ref{sec:passk}). The orchestrator's curve does not saturate over the same range: each additional orchestrator definer contributes a fresh trajectory through the spawn-augmented action space, and the resulting candidate pool is more diverse than the pipeline's at matched $M$. This is the surface signature of the orchestrator addressing the bottleneck the pipeline's diagnostic predicts.

\subsection{Generation-bound diagnostic and its architectural test}\label{sec:passk}

To characterize whether accuracy gains from $M>1$ sampling come from more diverse generation or from better selection, we apply the unbiased pass@$k$ estimator~\cite{chen2021codex} to the $M{=}5$ candidate sets of both the pipeline and the orchestrator.

\paragraph{Selection captures most of the candidate ceiling.}
On the pipeline's $M{=}5$ candidate distribution, unbiased pass@1 is 46.40\% (the accuracy a randomly chosen candidate would achieve in expectation), while naive pass@2 under train-score selection is 57.50\%. The pass@$M$ \emph{ceiling} --- the fraction of tasks for which at least one of five candidates is correct --- bounds what any selection policy can extract from the candidate pool; for the pipeline, selection recovers approximately 95\% of that ceiling. The same diagnostic on the orchestrator returns unbiased pass@1 of 56.21\% against naive pass@2 of 67.25\%, with selection again recovering $\sim$95\% of the candidate ceiling.

\paragraph{The system is generation-bound, not selection-bound.}
The per-task distribution of correct-candidate counts is strongly bimodal under both architectures: tasks tend to be either solved by most candidates or by none. Selection is already near-optimal; the hard ceiling on pass@2 at current $M$ is set by the fraction of tasks for which no candidate is correct. Improving headline accuracy therefore requires broadening the candidate distribution, through better exploration, larger $N$, or representational changes that alter the transformation space the synthesis stage searches, rather than improving selection fidelity among existing candidates.

\paragraph{The orchestrator as the architectural test.}
The diagnostic above predicts that an architecture which broadens candidate generation should produce a real lift, while one that improves selection fidelity should not. The orchestrator implements the former: spawn introduces fresh exploration trajectories conditioned on the failure modes of prior candidates, producing a meaningfully different candidate distribution per task. Unbiased pass@1 comparison confirms the prediction: pipeline unbiased pass@1 is 46.40\% and orchestrator unbiased pass@1 is 56.21\%, a $+9.81$\,pp lift that matches the $+9.75$\,pp naive pass@2 lift to within a tenth of a point. This near-equality is the strongest available signal that the orchestrator's lift is generation-side: were it selection-mediated, the naive pass@2 lift would substantially exceed the unbiased pass@1 lift. Hence, the orchestrator's lift offers confirmation that adaptive re-exploration is indeed a productive lever.

\subsection{Behavioral characterization of the orchestrator}\label{sec:behavior}

Two trace-level statistics characterize what the orchestrator actually does. First, an exit-mode distribution over the orchestrator's 1{,}983 definer invocations on the headline run: 65.8\% terminated via the structural exit (train-set accuracy reached 100\%), 25.8\% terminated via \texttt{MAX\_ITERATIONS}, 7.3\% via an agent-mediated \texttt{done} call, 1.1\% via no tool call, and $<$0.1\% via consecutive execution errors. The \texttt{MAX\_ITERATIONS} tail (averaging $\sim$4.6 attempts and $\sim$2.2 spawns per definer, all at train-score zero) is the dominant cost driver and motivates better orchestrator early-exit prompting. 

Second, spawn is empirically the dominant mechanism behind the orchestrator's lift. Of the 45 tasks on which the orchestrator solves and the pipeline does not, 31 (69\%) have a winning definer that called \texttt{explore\_new\_patterns}. Of the 40 tasks on which the orchestrator solves and \emph{all of} baseline, CoT, and pipeline fail, 30 (75\%) are spawn-induced. The architecture was designed around the hypothesis that wrong-abstraction failures are the dominant residual error mode and that spawn would be the relief valve; the cross-system task analysis confirms it. 

Appendix~\ref{sec:canonical} walks through one such task in detail: a geometric-versus-arithmetic rule trap in which both rules fit the training pairs perfectly. All pipeline definers commit to the arithmetic rule with no mechanism to revisit it. Under the orchestrator, a definer channels uncertainty into a focused spawn and recovers the geometric rule from the returned findings.
\subsection{Cross-model generality}\label{sec:crossmodel}

The architectural lifts reported above are measured on a single model (DeepSeek~V3.2). To test whether the pattern is V3.2-specific or general, we replicate the four-architecture comparison on Qwen3-235B-Instruct, a second open-weight model from a different model family (instruct variant; no thinking toggle). Due to budget constraints and rate-limiting on the Qwen3 provider, we run the cross-model comparison on a 99-task subset at $\{N{=}5, t{=}0.5, M{=}3\}$ for both the Explorer-Definer Pipeline and the Reflective Orchestrator --- $M{=}3$ is the largest ensemble depth Qwen3 could deliver under the rate-limit cap, so we match DeepSeek V3.2 to it for parity. The same 99 tasks and the same $M{=}3$ subsampling are applied to V3.2. All other components (harness, scoring, deduplication protocol, cost accounting) are held identical across models.

\begin{figure}[htbp]
    \centering
    \includegraphics[width=0.85\textwidth]{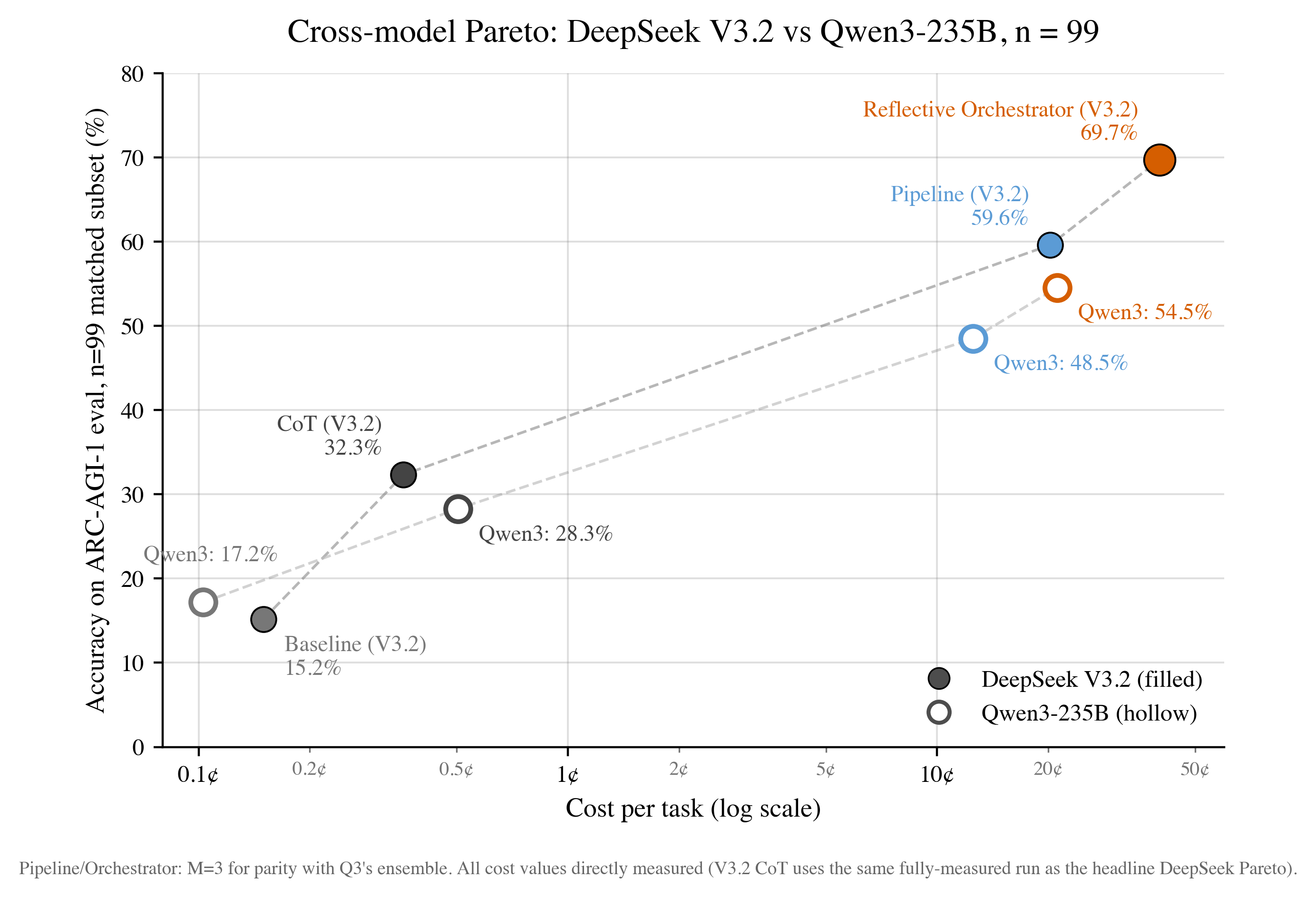}
    \caption{Cross-model Pareto on a matched 99-task subset of ARC-AGI-1. DeepSeek~V3.2 (filled markers) and Qwen3-235B-Instruct (hollow markers) traverse the same four-architecture progression. Pipeline and orchestrator use $M{=}3$ on both models for ensemble-depth parity. Cost axis is log-scale.}
    \label{fig:crossmodel}
\end{figure}

Figure~\ref{fig:crossmodel} plots the result. The architectural progression replicates on Qwen3: every lever produces a positive lift (baseline 17.17\% $\to$ CoT 28.28\% $\to$ pipeline 48.48\% $\to$ orchestrator 54.55\%), with the same monotone ordering as V3.2 on the same subset (15.15\% $\to$ 32.32\% $\to$ 59.60\% $\to$ 69.70\%).
\begin{table}[htbp]
\centering
\caption{Paired-bootstrap deltas on the matched 99-task subset at $M{=}3$. Within-model lifts compare each model's orchestrator to its own pipeline; cross-model gaps compare Qwen3 to V3.2 at matched architecture. CIs are 95\,\% paired bootstrap (B=10{,}000); ${}^*$ excludes zero.}
\label{tab:crossmodel_deltas}
\small
\begin{tabular}{lcc}
\toprule
Comparison & $\Delta$ (pp) & 95\,\% CI \\
\midrule
\multicolumn{3}{l}{\emph{Within-model: Pipeline $\to$ Reflective Orchestrator}} \\
\quad DeepSeek V3.2                         & $+10.10^*$ & $[+3.03,\; +17.17]$  \\
\quad Qwen3-235B                            & $+6.06$    & $[+0.00,\; +12.12]$  \\
\midrule
\multicolumn{3}{l}{\emph{Cross-model gap: Qwen3 $-$ V3.2 at matched architecture}} \\
\quad Pipeline                              & $-11.11^*$ & $[-18.18,\; -5.05]$  \\
\quad Reflective Orchestrator               & $-15.15^*$ & $[-23.23,\; -7.07]$  \\
\bottomrule
\end{tabular}
\end{table}

Table~\ref{tab:crossmodel_deltas} shows both pipeline $\to$ orchestrator lifts are directionally positive and consistent in rough magnitude with the full-evaluation V3.2 result ($+9.75$\,pp at $n=400$, Section~\ref{sec:paired}). While the Qwen3 lift CI does touch zero, it reflects the 99-task subset size rather than an architectural difference; the point estimate is comfortably positive but $n{=}99$ is not enough to exclude zero by paired bootstrap.

\paragraph{Interpretation.}
The cross-model result supports the paper's central claim that the harness, not the specific model, drives the lifts reported here. Though headline accuracy numbers are V3.2-specific, our three-lever architectural framing, the diagnostic-then-test evaluation arc, and the relief-valve role of spawn appear to generalize. Single-additional-model replication is not a definitive test of generality, and the $n{=}99$ subset prevents tight inference on the Qwen3 orchestrator lift specifically. Broader cross-family validation remains future work (Section~\ref{sec:futurework}).
\subsection{Cost--accuracy positioning}

\begin{figure}[htbp]
    \centering
    \includegraphics[width=0.85\textwidth]{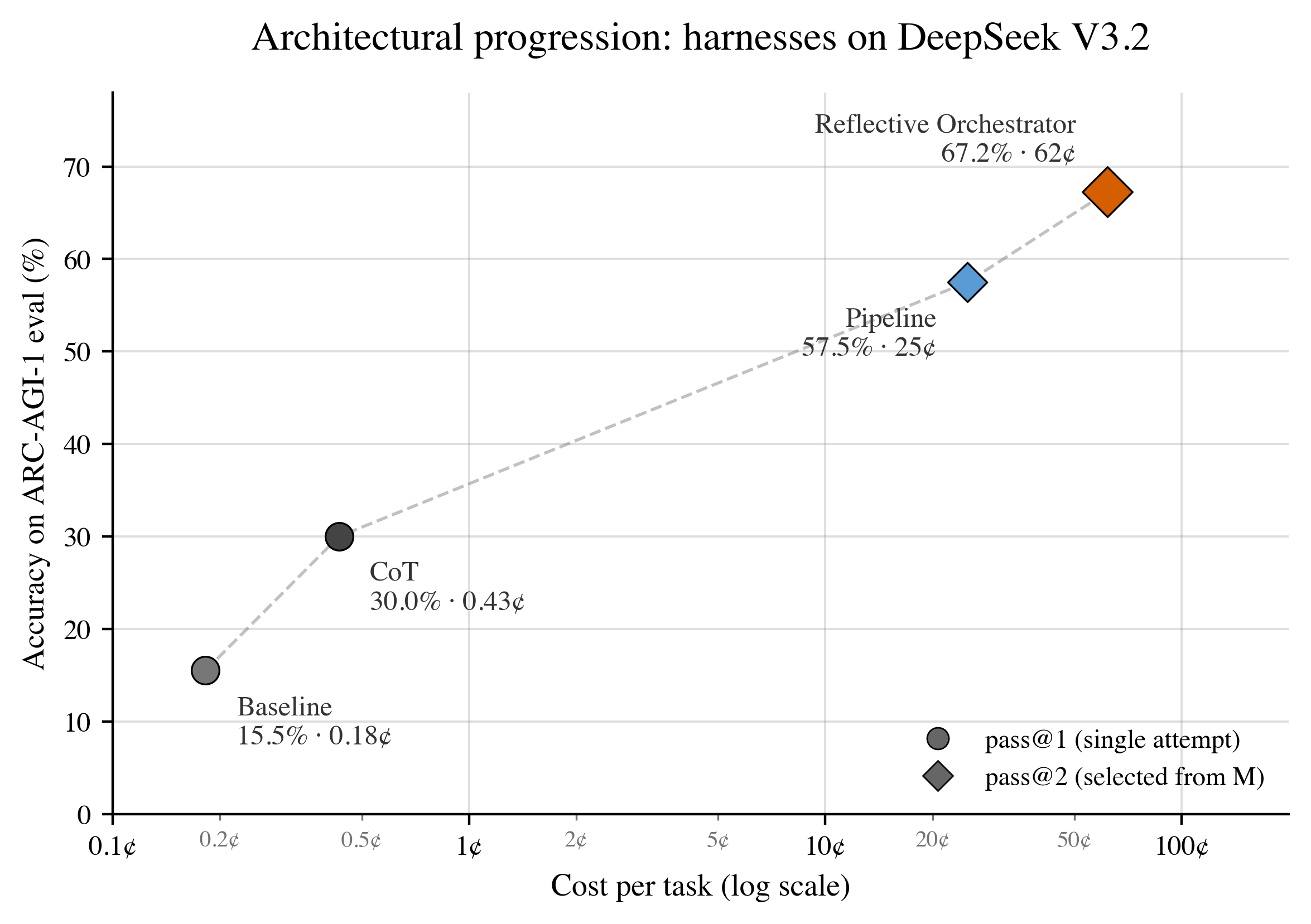}
    \caption{Headline cost--accuracy frontier across the four primary architectures. DeepSeek~V3.2 served via AtlasCloud FP8, ARC-AGI-1 public evaluation set ($n=400$). Cost axis is log-scale; cost at AtlasCloud FP8 list rates (\$0.30/M prompt, \$0.38/M completion). Each architectural lever (within-call CoT deliberation, across-call pipeline decomposition, and adaptive re-exploration in the orchestrator) corresponds to a monotone move along the frontier.}
    \label{fig:headline}
\end{figure}

Figure~\ref{fig:headline} places the four primary conditions on the cost--accuracy plane.

Within our evaluation regime, the pipeline at \$0.25 per task sits more than an order of magnitude below the cheapest published custom frontier-model system we have referenced at comparable accuracy (Pang at $\sim$\$2.56 per task and 77.1\% on ARC-AGI-1~\cite{pang2025}), and approximately two orders of magnitude below Berman's 2024 evolutionary system at 53.6\% and $\sim$\$29 per task~\cite{berman2024}. The orchestrator at \$0.62 per task extends the frontier upward by nearly ten points without moving out of the sub-\$1 regime, and remains roughly $4\times$ cheaper than Pang and $\sim$45$\times$ cheaper than Berman's 2024 system at comparable or higher accuracy. As stated in Section~\ref{sec:limitations}, these comparisons are contextual: published systems are evaluated under slightly different protocols (e.g., reporting semi-private or competition-set results), so cost contrast should be read as a regime-level observation rather than a head-to-head benchmark.

%----------------------------------------------------------------------------------------
%	5. DISCUSSION
%----------------------------------------------------------------------------------------

\section{Discussion}\label{sec:discussion}

\subsection{Three architectural levers, sequentially compounding}

The clearest signal in our results is that three distinct architectural levers each contribute meaningfully and compound sequentially when stacked. Adding a structured reasoning channel to a single LLM call (CoT) lifts accuracy by $+14.50$\,pp over an unscaffolded baseline. Adding the Explorer-Definer Pipeline on top lifts it by a further $+27.50$\,pp pass@2. Adding adaptive re-exploration (the Reflective Orchestrator) on top of the pipeline lifts it by a further $+9.75$\,pp pass@2. The total architectural lift on the same model is 52 points, with no benchmark-specific training or heavy test-time compute.

These three lifts are unlikely to be redundant. CoT externalizes within-call reasoning. The pipeline introduces structural decomposition --- different stages, with compressed natural-language artifacts between them and an executable verifier at the back end --- that a single-call architecture cannot easily replicate. The orchestrator addresses a different failure mode again: not insufficient reasoning per call (CoT's lever) and not insufficient structure across calls (pipeline's lever), but rigidity in the upstream-downstream boundary. That is, the inability of a synthesis stage to route back to pattern discovery when its hypothesis is wrong. Each lever is the architectural response to a different bottleneck.

This is consistent with, but distinct from, the test-time compute scaling literature. Snell et al.~\cite{snell2024scaling} show that sampling more from a smaller model can outperform a single pass from a larger one; our result is that \emph{how} additional calls are organized matters as much as their number. Each of our three architectures spends more model calls than the previous one, but it spends them on structurally different things: externalized reasoning, role specialization, or adaptive re-exploration, and each is spent on the bottleneck the previous architecture left in place.

\subsection{The \texttt{think} tool is a measurable component}

The act-only ablation gives a controlled measurement of an architectural detail that is often an obscure prompting-layer component. Disabling the definer's \texttt{think} scratchpad reduces pass@2 from 57.50\% to 51.75\% (\emph{while reducing cost by 32\%}); the paired-bootstrap CI on the $-5.75$\,pp delta excludes zero comfortably. The remaining 51.75\% still sits more than 20\,pp above the CoT baseline (30.00\%) and more than 35\,pp above the unscaffolded baseline (15.50\%), so the bulk of the pipeline's lift over single-call architectures comes from agentic decomposition, role specialization, and executable verification rather than from the scratchpad alone. Within the pipeline architecture, though, the \texttt{think} channel is doing measurable work that the structured tool fields available in the act-only configuration cannot substitute for.

This ablation does not separate whether \texttt{think}'s contribution comes from \emph{additional} deliberation tokens, from the \emph{structure} of a private scratchpad held separate from the output channels, or from both at once. A token-budget reading (more deliberation tokens buy better accuracy) is consistent with what we observe, and is the same lever the CoT baseline exhibits ($+14.50$\,pp over the unscaffolded baseline at $\sim$2$\times$ cost). What the ablation does establish is that tool-mediated scratchpads are a meaningful architectural lever on top of agentic decomposition, not a stylistic detail that can be dropped without consequence.

\subsection{A diagnostic that produced a falsifiable prediction and was confirmed}

The unbiased pass@$k$ analysis on the pipeline is a methodological centerpiece. It produced the following falsifiable prediction: because selection already captures $\sim$95\% of the candidate ceiling, the binding constraint is candidate generation, and any architecture that broadens generation should produce a real lift, while any architecture that improves selection should not. The orchestrator is the design that follows the prediction's logic: it does not improve ranking among existing candidates; it changes which candidates exist, by allowing the synthesis stage to spawn fresh exploration when its current trajectory fails. The confirming evidence is the near-equality of the orchestrator's lift on the two estimators: $+9.75$\,pp on naive pass@2 (selection-mediated) and $+9.81$\,pp on unbiased pass@1 (selection-free). Were the orchestrator's lift driven by easier selection on an unchanged candidate distribution, the unbiased lift would have been substantially smaller than the naive lift. It is not.

This pattern --- diagnose the bottleneck, design the architecture to target it, verify with the same diagnostic from the opposite end --- is the discipline we would advocate for compute-efficiency claims in agentic systems more generally. Headline accuracy lifts are straightforward to produce by spending more on inference; what is harder is to spend that compute on the right axis. The diagnostic-then-target loop can signal which axis is ``correct'' before one spends.

\subsection{Spawn as the relief valve}

The orchestrator's architecture was designed around a specific hypothesis: that the pipeline's residual error is dominated by wrong-abstraction failures, in which the explorer stage's initial pattern findings miss the operative rule, and the definer (constrained to refine on those findings) cannot recover. The cross-system task analysis (Section~\ref{sec:behavior}) corroborates the hypothesis: 75\% of the tasks the orchestrator solves and all three baselines fail are tasks on which the winning orchestrator definer called \texttt{explore\_new\_patterns}. The canonical example in Appendix~\ref{sec:canonical} is a concrete instance of the mechanism: a task on which an arithmetic rule fits the training pairs but fails the test, four of five orchestrator definers (and all five pipeline definers) lock onto the arithmetic rule, and only the spawning definer escapes the wrong frame to discover the underlying geometric rule. This is not a one-off: the trace pattern of multiple definers converging on a wrong rule that fits training pairs, while one dissenter recovers via mid-loop spawn, recurs across the 30 spawn-induced unique solves the orchestrator finds.

The dedup-by-test-prediction selection mechanism is also a highly significant component: without it, the four convergent wrong candidates in the above example would have dominated the pass@2 submission and the spawning definer's correct candidate would have been crowded out. Selection here functions as implicit diverse voting, treating $n$ convergent definers as one vote and amplifying the dissenter --- a design choice that pairs naturally with mid-loop spawn, since spawn is an architectural channel that is intended to transform uncertain definers into potential dissenters via re-exploration.

\subsection{Cost--accuracy positioning}

The pipeline at \$0.25 per task and the orchestrator at \$0.62 per task together occupy a region of the cost--accuracy plane that published systems have largely skipped. At 57.50\% pass@2, the pipeline is comparable in absolute accuracy to Berman's 2024 evolutionary system (53.6\% at $\sim$\$29 per task) at roughly two orders of magnitude lower cost~\cite{berman2024}, and is roughly an order of magnitude cheaper than the closest efficiency-oriented prior work --- Pang's program-synthesis system~\cite{pang2025}, which holds the efficiency frontier \emph{within} the frontier-model search regime at $\sim$\$2.56 per task and 77.1\%. The orchestrator at 67.25\% extends the frontier upward by nearly ten points while staying under \$1 per task. We do not approach the absolute accuracy ceiling set by Berman's 2025 natural-language system (79.6\% at \$8.42 per task)~\cite{berman2025} or by other competition-grade systems. The contribution is not a new state of the art in accuracy; it is evidence that a substantial fraction of ARC-AGI-1 performance is recoverable through architecture alone, on a general-purpose open-weight model, with no benchmark-specific training or costly compute at inference time. The comparison is regime-level rather than head-to-head: we emphasize that external systems are evaluated under different protocols, and the cost contrast should be read as a frontier-position statement rather than a universal benchmark.

%----------------------------------------------------------------------------------------
%	6. LIMITATIONS
%----------------------------------------------------------------------------------------

\section{Limitations}\label{sec:limitations}

\paragraph{Public evaluation set, not semi-private leaderboard.}
This study evaluates on the ARC-AGI-1 public 400-task evaluation set because the semi-private leaderboard service is not available for arbitrary local architecture experiments~\cite{arcprize_policy,arcprize_arcagi1}. We therefore do not report a verified leaderboard score, and our results should be read as controlled within-model architectural comparisons rather than verified leaderboard submissions. External system numbers (Berman, Pang, Greenblatt, TRM) are reported only as contextual reference points for the cost--accuracy regime; matched-protocol comparison is not possible without semi-private access. A consequence is that public-set results may overestimate semi-private generalization, particularly if any of the architectures inadvertently exploit public-set artifacts. The same harnesses, model, prompts, and scoring code are used across all conditions, so any shared public-set bias is held constant across the within-paper comparisons; our architectural \emph{deltas} therefore remain interpretable even if the absolute level shifts.

\paragraph{Configuration selection on the evaluation set.}
The canonical pipeline configuration ($N{=}5,\,t{=}0.5,\,M{=}5$) was selected by inspecting the $N \times t \times M$ Pareto surface (Section~\ref{sec:surface}) on the same 400-task public evaluation set used for downstream comparisons, rather than on a held-out validation split. A methodologically cleaner protocol would have selected the configuration on a validation subset and evaluated on the remainder. Three factors mitigate this in our specific case. First, the sweep surface is smooth: configurations near the canonical perform within a few percentage points, so the headline numbers do not depend on a sharp selected point. Second, the architectural deltas (baseline $\to$ CoT, CoT $\to$ pipeline, pipeline $\to$ orchestrator) hold across most of the surface, so the central claim that architecture drives the lift is not a single-config artifact. Third, the cross-model replication on Qwen3-235B (Section~\ref{sec:crossmodel}) preserves the same architectural ordering at the same canonical configuration on a different model family, which would not be the expected outcome if the configuration were overfit to V3.2-specific evaluation-set quirks. Reported paired bootstrap confidence intervals on this single 400-task set may underestimate true population variance.

\paragraph{Single model family at full scale.}
Headline results use DeepSeek~V3.2 served via AtlasCloud FP8. The model has undergone general-purpose post-training (Section~\ref{sec:related}); that training is held constant across architectures, so it cannot confound the within-model architectural deltas, but it does set the absolute level. Our initial cross-model validation on Qwen3-235B (Section~\ref{sec:crossmodel}) finds that the four-architecture progression and the pipeline $\to$ orchestrator lift both replicate in direction and rough magnitude; the absolute level is model-specific (Qwen3 trails V3.2 by 11--15\,pp at matched architecture on this subset). The central claim, that architectural decomposition and adaptive re-exploration recover much of the compute--accuracy gap, would be stronger if it held across additional model families and at full-evaluation scale; the Qwen3 cross-model subset was constrained by provider rate limits to 99 tasks, which prevents tight inference on the Qwen3 orchestrator lift specifically.

\paragraph{Run-to-run variance.}
Accuracy figures are point estimates from single runs. Bootstrap CIs capture sampling uncertainty over the 400-task set but not execution noise from non-determinism in the inference stack. Informal cross-ablation comparison suggests a $\sim\pm2$\,pp single-run noise floor. Explicit characterization with repeated runs of the headline cell remains future work, and is particularly relevant for the smallest reported deltas (the $-0.75$\,pp refinement ablation and the $+2.75$\,pp pass@1 $\to$ pass@2 bonus). The orchestrator's adaptive exit conditions also introduce per-task wall-clock variance absent in the pipeline, though dollar cost is token-based and thus unaffected.

\paragraph{ARC-AGI-1 only.}
ARC-AGI-2 tasks are designed to resist single-step pattern matching and stress multi-step compositional reasoning~\cite{arcagi2}. Whether the lifts reported here survive on harder material, or whether they partly reflect a memorization floor shared by all conditions, remains to be settled by running the same harnesses on ARC-AGI-2.

%----------------------------------------------------------------------------------------
%	7. FUTURE WORK
%----------------------------------------------------------------------------------------

\section{Future Work}\label{sec:futurework}

The generation-bound diagnostic that motivated the Reflective Orchestrator points to several extensions of the same logic: continue investing upstream in candidate diversity, with the orchestrator's spawn mechanism as the starting point rather than the endpoint.

\paragraph{Spawn-off ablation.}
The cleanest causal isolation of spawn's contribution is to rerun the orchestrator with \texttt{explore\_new\_patterns} disabled and observe the residual lift from the agentic loop structure alone. The cross-system task analysis (Section~\ref{sec:behavior}) gives strong correlational evidence that spawn drives the majority of orchestrator-unique solves, but a dedicated ablation run would convert that into a paired delta. This was deferred from the present study only for cost reasons.

\paragraph{Spawn fan-out and depth.}
The orchestrator fixes $K{=}2$ explorers per spawn and caps spawn at three calls per task, both motivated by pipeline-derived heuristics rather than orchestrator-specific sweeps. A targeted $K \times \text{spawn-cap}$ sweep would characterize whether the orchestrator's own optimum differs, and whether deeper nested re-exploration (a spawned explorer's findings triggering a further or broader spawn) is productive or wasteful.

\paragraph{Learned or signal-driven branch policy.}
Within the orchestrator architecture, the definer agent's branch decision (e.g., continue refining, spawn, or commit) is model-mediated, driven from the structured train-pair diffs. Though less ``agentic,'' a learned branch classifier, or a self-certainty signal derived from Kang et al.~\cite{kang2025selfcertainty}, could route earlier and more reliably. The orchestrator's exit-mode distribution (Section~\ref{sec:behavior}) shows that the \texttt{MAX\_ITERATIONS} tail is the dominant cost driver, so a sharper branch policy through prompting alone, heuristics, or other means could improve both accuracy and cost.

\paragraph{Self-certainty-weighted explorer ranking.}
A complementary upstream lever is to rank explorer outputs by per-explorer self-certainty before passing them to the synthesis stage, rather than concatenating them in agent-index order (effectively with uniform weight). Kang et al.~\cite{kang2025selfcertainty} propose self-certainty: a per-generation confidence estimate computed from token-level logprobs as the average KL divergence between each token's output distribution and the uniform vocabulary distribution. Ranking explorers via Borda voting on those scores would let the definer attend preferentially to high-confidence pattern documents, on the hypothesis that confident explorers surface more reliable abstractions and that boosting their salience reduces the rate at which the definer commits to low-confidence (and often wrong-frame) hypotheses from the fleet. We treat this as future work for the practical reason that the endpoint we use does not currently expose token-level logprobs. The intervention becomes feasible on any logprob-supported provider or model and pairs naturally with the orchestrator: a more confidently-ranked explorer fleet plausibly reduces the rate of orchestrator \texttt{explore\_new\_patterns} calls, shifting cost away from the significant \texttt{MAX\_ITERATIONS} tail.

\paragraph{Cross-model panel extension.}
The Qwen3-235B cross-model run (Section~\ref{sec:crossmodel}) is currently constrained to a 99-task subset by provider rate limits. Extending it to the full ARC-AGI-1 evaluation set would tighten the Qwen3 pipeline $\to$ orchestrator paired CI, which currently touches zero at $n=99$. Beyond that, a third model family --- particularly a non-MoE architecture (e.g., Llama-3.3-70B) or a smaller-tier instruct model --- would test whether the generality observation holds across capability tiers and architectural classes.

\paragraph{ARC-AGI-2.}
Running the same harness on ARC-AGI-2~\cite{arcagi2} would test whether the lifts survive on tasks designed to resist single-step pattern matching, and would partially address public-set contamination concerns by evaluating on fresher material.

\paragraph{Direct-grid vs.\ code-output ablation.}
The pipeline's and the orchestrator's choice to require executable Python rather than direct grid prediction is structurally important but has not been ablated. A controlled comparison with the same explorers, and compressed artifacts but with a synthesis stage that emits a grid directly would isolate the contribution of executable-code output and the verifier signal it enables. This was avoided for our scope because direct-grid generation eliminates easily attainable training set accuracy, which complicates the current selection mechanism implementation.

\paragraph{Hard-tail failure taxonomy.}
A qualitative classification of the tasks the orchestrator still cannot solve would sharpen the generation-bound claim further: distinguishing exploration misses that even spawn cannot recover, from synthesis misses (the right pattern is identified but the orchestrator fails to program it), would inform promising levers to investigate.

\paragraph{Multi-modal exploration.}
Providing explorers with rendered images of the grids alongside the symbolic representation is a plausible upstream lever: visual gestalt features (symmetry, containment, alignment) may be more readily extracted from pixel-level input than from tokenized arrays. This is a speculative extension and is not evaluated here.

%----------------------------------------------------------------------------------------
%	REFERENCES
%----------------------------------------------------------------------------------------

\printbibliography

%----------------------------------------------------------------------------------------
%	APPENDIX
%----------------------------------------------------------------------------------------

\appendix

\section{Canonical task: \texttt{3a301edc}}\label{sec:canonical}

This appendix walks through a single ARC-AGI-1 evaluation task on which the Reflective Orchestrator's spawn mechanism is decisive. The task is failed by the baseline, by CoT, and by the pipeline; the orchestrator solves it after one of its five definers called \texttt{explore\_new\_patterns}. The trace illustrates the mechanism in concrete terms and exposes a subtler property of the task: the training pairs do not just \emph{underdetermine} the correct rule, they are \emph{equivalently described} by two distinct rule families, both of which produce identical predictions on all five training outputs.

\subsection{The task}

\begin{figure}[htbp]
    \centering
     \includegraphics[width=0.85\textwidth]{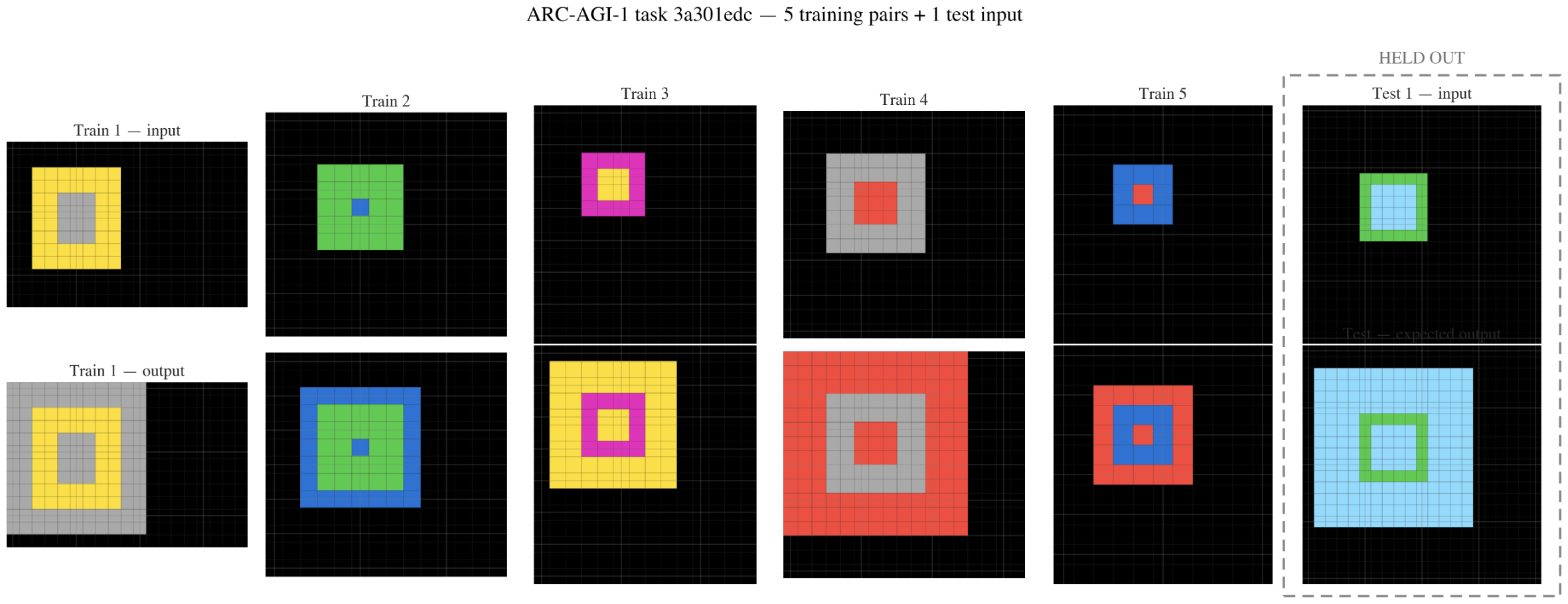}
    \caption{Task \texttt{3a301edc}, training pairs and test pair (input on top, expected output on bottom).}
    \label{fig:canonical_task}
\end{figure}

\paragraph{The training pairs cannot distinguish the rules.}
Table~\ref{tab:canonical_predictions} evaluates both rules on every training pair. On all five, the two rules produce identical predictions; their disagreement only manifests on the test pair.

\begin{table}[htbp]
\centering
\caption{Predicted border thickness under each rule for every pair in task \texttt{3a301edc}. The two rules agree on all five training pairs and disagree only on the test pair, where the arithmetic rule overestimates by a single cell. Gap and color values per row are read off the inputs in Figure~\ref{fig:canonical_task}; the arithmetic-rule column shows the resulting $\max(\text{gap}, |\Delta\text{color}|)$ computation.}
\label{tab:canonical_predictions}
\small
\begin{tabular}{lcccc}
\toprule
Pair & Inner shape & Geometric rule & Arithmetic rule & Actual \\
\midrule
Train 1 & $4 \times 3$ (rectangle) & $\min(4,3) - 1 = 2$ & $\max(2, 1) = 2$  & 2 \\
Train 2 & $1 \times 1$ (single)    & side $= 1$          & special-case $\to 1$ & 1 \\
Train 3 & $2 \times 2$ (square)    & side $= 2$          & $\max(1, 2) = 2$  & 2 \\
Train 4 & $3 \times 3$ (square)    & side $= 3$          & $\max(2, 3) = 3$  & 3 \\
Train 5 & $1 \times 1$ (single)    & side $= 1$          & special-case $\to 1$ & 1 \\
\midrule
\textbf{Test} & $\bm{4 \times 4}$ \textbf{(square)} & $\bm{\text{side} = 4}$ & $\bm{\max(1, 5) = 5}$ & \textbf{4} \\
\bottomrule
\end{tabular}
\end{table}

In the pipeline, no amount of refinement on the arithmetic rule from training feedback alone can cause preference of another rule, so both reach $\text{train} = 100\%$ by construction. The disambiguation requires a prior over rule families: a definer either recognizes the arithmetic formula as an ad-hoc explanation that needs structural support, or it does not.

\subsection{Pipeline behavior: the off-by-one failure}

All five pipeline definers on this task reach $\text{train} = 100\%$ on the arithmetic rule. Their predictions on the held-out test pair are identical: a new outer border of thickness \emph{five}, anchored five cells outside the original outer region in every direction. The expected test output has a border of thickness four. The pipeline is wrong by a single cell on every side, and identically so across all five definers.

Dedup-by-test-prediction collapses the five convergent definers into a single candidate, so the pass@2 submission contains exactly one distinct candidate (the wrong one) and the task fails. This is the failure mode the generation-bound diagnostic predicts at the trace level: selection is not the problem (all five candidates being wrong in the same way is exactly the case in which no selection policy over the candidate pool can succeed), and the problem is not training-feedback either (all five are at $\text{train} = 100\%$). The problem is that the candidate distribution does not contain the geometric rule. The pipeline's directed handoff prevents the definer from returning to exploration once it has committed to the arithmetic frame.

\subsection{Orchestrator behavior: how spawn unlocks the right frame}

Four of five orchestrator definers also lock onto the arithmetic rule and reach $\text{train} = 100\%$ with it. The fifth definer, after the same initial confidence, becomes structurally suspicious of its own hypothesis; unlike in the pipeline, it is equipped to call \texttt{explore\_new\_patterns}, doing so with primarily \emph{negative} guidance:

\begin{quote}
\small
``The PatternExplorers have conflicting hypotheses. Border color always equals innermost region's color. Border thickness $=$ |outer-inner| fails Examples 1,2. Border thickness $=$ gap size: gap is 1 for all examples but thickness varies. Border thickness $=$ border color value: fails Example 4 (color=2 but thickness=3). Positive hypotheses to explore: is thickness $=$ \texttt{min(inner\_width, inner\_height)}? Does the algorithm respect the inner region's shape (square vs.\ rectangular)?''
\end{quote}

The definer's specific arithmetic claims in this guidance are imprecise, since the gap is in fact $2, 2, 1, 2, 1$ across the five training pairs rather than $1$ throughout, but the structural conclusion holds: thickness does not track gap in a clean functional way (gap $= 2$ in Train 1 with thickness $2$ but gap $= 2$ in Train 2 with thickness $1$), so a gap-only formula cannot be the rule. What matters is the agent's expressed shift in category: from arithmetic-feature explanations toward shape-based ones. The single positive hypothesis at the end of the guidance --- ``Does the algorithm respect the inner region's shape (square vs.\ rectangular)?'' --- is the critical redirect.

The two spawned focused explorers, conditioned on this guidance, iteratively examine the described pattern and return findings that converge on the geometric rule. In particular, a returned PatternExplorer states explicitly: 

\begin{quote}
\small
``If innermost region is square (width $=$ height), border thickness equals width (or height) of that square region. If innermost region is rectangular (width $\neq$ height), border thickness equals $\min(\text{width}, \text{height}) - 1$.'' 
\end{quote}

The orchestrator definer reads this, redefines the transformation from scratch using a shape-conditional branch, and reaches $\text{train} = 100\%$ on a categorically different program from the other four definers.

The substantive enabler is not that the spawning definer was more careful in its reasoning (we note above that its trace contains arithmetic mistakes the other four do not). Rather, the enabler is that the architecture exposes a tool (\texttt{explore\_new\_patterns}) that lets one definer redirect the search toward a different category of explanation, conditioned on a brief structural intuition that the current category is wrong. The spawn mechanism turns a vague structural suspicion about the arithmetic formula into a concrete re-exploration with shape-oriented guidance, which is enough for the spawned explorers to gather evidence and surface the correct rule.

\subsection{Why selection picks the winner}

All five orchestrator definers reach $\text{train} = 100\%$, so the training-set score alone is uninformative. Without dedup-by-test-prediction, the top-2 pass@2 candidates would be the first two (arithmetic-rule) definers by agent index and the task would fail. With dedup-by-test-prediction, the four convergent arithmetic-rule definers merge into a single candidate and the spawning definer's geometric-rule prediction becomes a second distinct candidate; both are submitted as pass@2 and the geometric prediction matches the expected test output exactly.

The mechanism is selection-architectural in a specific sense: dedup-by-test-prediction is the selection rule that lets a single dissenter outvote a convergent majority. It pairs naturally with mid-loop spawn because spawn is the mechanism that produces dissenters in the first place. Together they encode the assumption that on hard ARC tasks where multiple rules fit the training pairs, the productive search direction is to amplify candidate diversity, not to compound candidate confidence. Task \texttt{3a301edc} is one instance of this pattern; the cross-system analysis (Section~\ref{sec:behavior}) shows that approximately 75\% of orchestrator-unique solves follow the same trace shape.

\end{document}